\documentclass[twoside,11pt]{article}

\usepackage{jmlr2e}
\usepackage[colorinlistoftodos]{todonotes}
\usepackage{nomencl}
\usepackage[T1]{fontenc}
\usepackage[utf8]{inputenc}
\usepackage{tabularx,ragged2e,booktabs,caption}
\usepackage{todonotes}
\usepackage{amsmath}
\usepackage{tcolorbox}
\usepackage{array}
\usepackage{bbm}
\usepackage{color, colortbl, hhline}
\usepackage{multirow}
\usepackage{arydshln}
\newcolumntype{C}[1]{>{\centering\arraybackslash}m{#1}}

\usepackage{tikz}
\usepackage{subfig}
\usepackage{floatrow}
\usetikzlibrary{arrows,automata,positioning}

\usepackage{algpseudocode,algorithm,algorithmicx}

\DeclareMathOperator*{\argmax}{arg\,max}

\newcommand{\R}{\mathbb{R}}

\newcommand{\E}{\mathbb{E}}

\renewcommand{\P}{\mathbb{P}}

\newcommand{\norm}[1]{\| #1 \|}

\newcommand{\states}{\mathcal{S}}
\newcommand{\actions}{\mathcal{A}}

\newcommand{\Rmax}{R_{\mathrm{max}}}

\newcommand{\s}{\mathcal{S}}
\renewcommand{\a}{\mathcal{A}}

\renewcommand{\r}{\mathcal{R}}
\newcommand{\p}{\mathcal{P}}
\newcommand{\x}{\mathcal{X}}

\let\vec\boldvec

\makenomenclature

\ShortHeadings{A Survey of Exploration Methods in Reinforcement Learning}{xxxx}
\firstpageno{1}

\begin{document}

\title{A Survey of Exploration Methods in Reinforcement Learning}


\author{\name Susan Amin \email susan.amin@mail.mcgill.ca \\
       \addr Department of Computer Science, McGill University\\
  Mila- Québec Artificial Intelligence Institute\\
       Montréal, Québec, Canada\\
       \AND
       \name Maziar Gomrokchi \thanks{These authors contributed equally to the work} \email gomrokma@mila.quebec \\
       \addr Department of Computer Science, McGill University\\
  Mila- Québec Artificial Intelligence Institute\\
       Montréal, Québec, Canada\\
       \AND
       \name Harsh Satija \footnotemark[1]\email harsh.satija@mail.mcgill.ca \\
       \addr Department of Computer Science, McGill University\\
  Mila- Québec Artificial Intelligence Institute\\
       Montréal, Québec, Canada\\
       \AND
       \name Herke van Hoof \footnotemark[1]\email h.c.vanhoof@uva.nl \\
       \addr Informatics Institute, University of Amsterdam\\
       Amsterdam, the Netherlands\\
       \AND
       \name Doina Precup \email dprecup@cs.mcgill.ca \\
       \addr Department of Computer Science, McGill University\\
  Mila- Québec Artificial Intelligence Institute\\
       Montréal, Québec, Canada}

\editor{xxx}

\maketitle

\begin{abstract}
Exploration is an essential component of reinforcement learning algorithms, where agents need to learn how to predict and control unknown and often stochastic environments. Reinforcement learning agents depend crucially on exploration to obtain informative data for the learning process as the lack of enough information could hinder effective learning. In this article, we provide a survey of modern exploration methods in (Sequential) reinforcement learning, as well as a taxonomy of exploration methods.
		\end{abstract}

\section{Introduction}

When a reinforcement learning (RL) agent starts acting in an environment, it usually does not have any prior knowledge regarding the task which it needs to tackle. The agent must interact with the environment, by taking actions and observing their consequences (in the form of rewards and next states), and then it can use this data to improve its behavior, as measured by the expected long-term return. This reliance on data that it gathers by itself differentiates RL agents from those performing either supervised or unsupervised learning, and it is often a limiting factor in terms of the agent's ultimate success at mastering the environment.
Specifically, if the agent only manages to visit a limited portion of the environment, its knowledge will be limited, leading to sub-optimal decision making \citep{wiering1999explorations}. However, if the agent focuses on acquiring information regarding  parts of the environment that it has not seen sufficiently,  it can lose the chance of gaining immediate reinforcement. This problem is referred to as \emph{exploration-exploitation} trade-off, and is a crucial open problem in  reinforcement learning (alongside generalization). Handling this trad-eoff is influenced by several factors, including the dynamics of the environment (\emph{i.e.} transition probability and reward distribution), properties of the state/action spaces (\emph{e.g.} discrete/continuous, number of states/actions, \dots), and the available number of interactions with the environment that the agent is allowed while it is training.

Our goal in this survey is to provide a broad high-level overview on the types of exploration methods employed by RL agents, by reviewing literature from the last three decades.   
Exploration studies have evolved during this time from simple ideas such as pure randomization, to increasingly effective methods which have interesting theoretical guarantees, or impressive empirical performance in large problems.

Exploration techniques have been categorized generally into \emph{undirected} and \emph{directed} methods \citep{thrun1992efficient} based on the choice of  information considered by the exploration algorithm. While in undirected exploration strategies, reinforcement learning agents select exploratory actions at random, without using \emph{any  exploration-specific knowledge}, directed exploration methods use the obtained information to pursue the exploration of less- visited state-action pairs, or of state-action pairs that are deemed to be more informative for the agent. 
However, this is not the only relevant differentiation between current exploration methods, as the field has expanded drastically and more nuances have developed.

In this survey, we present a more detailed categorization of exploration methods in reinforcement learning (see Figure~\ref{fig:categories}). We attempt to group existing algorithmic approaches into these categories in order to provide a more detailed understanding of the current landscape of methods, in terms of both the goals and the information they employ. Of course, a few of the methods do not fall squarely into one category, and hence are included in multiple categories.

We note that some of the discussed techniques in this survey were originally proposed and designed for bandit settings, and only later applied in the reinforcement learning problems. However, this survey is not intended to cover exploration methods specifically designed for bandits, as many references exist in this area, including the excellent recent textbook~\cite{lattimore2020bandit}. We will focus only on sequential decision making, where exploration has an even larger impact, as it controls not only the immediate information received by the agent, but also the potential interesting information in its future data stream. 
In addition, the very large number of publications in this field, especially in recent years, requires making hard choices regarding which papers to discuss in this survey. In some cases, we have picked particular representative methods for certain approaches, rather than listing all instances of a particular approach. Finally,  in order to provide a concise and easy-to-understand overview of the categories and methods, we do not focus on the mathematical details and theoretical results in the field. Our main goal is to provide an entry point into the field, which is accessible to readers who may want to understand the types of algorithms and empirical evaluations that have been provided, and to practitioners who want to build successful applications, and hence have to tackle this really difficult problem. 

The survey is organized as follows. In the following section, the notation is introduced and a brief RL background is provided. The exploration categories are subsequently presented in Section~\ref{overview}. Section~\ref{sec:pure} presents exploration methods that do not use reward information at all. Section~\ref{sec:stochastic} presents methods that rely mainly on randomizing action choices. Section~\ref{Bonus} presents methods that are based on optimism in the face of uncertainty. Section~\ref{sec:prior} introduces methods which start with the formulation of the optimal exploration-exploitation trade-off, and then approximate its solution. Section~\ref{sec:thompson} discusses probability matching methods, including posterior sampling. Finally, Section~\ref{conclusion} provides some conclusions and perspectives.

\section{Notation and Background}
In reinforcement learning problems, at each discrete time step $t=0,1,2,\dots$, the system is at state $s_t$ and the agent interacts with the environment by selecting action $a_t$. Consequently, the system transitions from the state $s_t$ to $s_{t+1}$, determined by the transition model of the system $\mathcal{P}$ and returns a numerical reward $r_t$. Upon receiving the reward, the agent decides on the next action $a_{t+1}$ and the process continues. Depending on the type of the problem, the decision making process either stops when it reaches a terminal state in an \textit{episodic task} or continues in a \textit{continuing task} with an infinite horizon. In the following paragraphs, we introduce the commonly used notation in this survey. Note that capital and calligraphic letters denote random variables and sets, respectively, unless stated otherwise. For a comprehensive introduction to reinforcement learning, refer to \citet{sutton1998introduction}.

\subsection{Markov Decision Processes}
In this survey, the problems are modeled as Markov decision processes (MDPs) $\mathcal{M}=\langle\states, \actions, \mathcal{P}, r\rangle$, where $\states$ and $\actions$ denote sets of all possible states and actions in the system, respectively. MDPs assume that the environment is Markovian; \emph{i.e.} the transition probability distribution $\mathcal{P}:\states\times\actions\rightarrow\mathcal{D}\left(\states\right)$ determines the next state from the probability distribution over the set of states $\mathcal{D}\left(\states\right)$ as a function of the \emph{current} state-action pair only. In other words, in MDPs we have,
\begin{equation}
    \P\left(S_{t+1}=s^\prime\mid s_t,a_t\right)=\P\left(S_{t+1} = s^\prime\mid s_t,a_t,s_{t-1},a_{t-1},\dots,s_0,a_0\right).
\end{equation}
If the agent starts from state $s$ and takes an action $a$, it transitions to the state $s^\prime$ with the probability,
\begin{equation}
    P\left(s,a,s^\prime\right) =\P\left(S_{t+1}=s^\prime\mid S_t=s,A_t=a\right).\label{eq:transitionProbability}
\end{equation}
The expected rewards can be written as the reward function $r:\states\times\actions\rightarrow\R$, which maps the current state-action pair $(s,a)$ to the immediate reward obtained from the set of real numbers $\R$ in the system,
\begin{equation}
    r\left(s,a\right)=\mathbb{E}\left[R_{t+1}\mid S_t=s,A_t=a \right],\label{eq:rewardFunction}
\end{equation}
where the expectation is with respect to the randomness induced by the reward function $r$.

All the sequence of observations, actions and any kind of information the agent obtains during its lifetime is called history,
\begin{equation}
    H_t=S_0,A_0,R_0,S_1,A_1,R_1,\dots,S_t,A_t,R_t.
\end{equation}

A trajectory $\tau$ is defined as the sequence of information extracted from the history $H_T$ for the horizon $T$.

\subsection{Reinforcement Learning Setting}
In reinforcement learning setting, agent behaves according to a policy $\pi\in\Pi$, where $\Pi$ represents the set of all possible policies. A deterministic policy $\pi:\states\rightarrow\actions$ at time step $t$, also represented as $a_t=\pi\left(s_t\right)$, returns a particular action $a_t$, while a stochastic policy $\pi\left(a_t\mid s_t\right)$ gives a probability distribution over a set of actions $\left(\pi:\states\rightarrow\mathcal{D}\left(\actions\right)\right)$, defined as
 \begin{equation}
     \pi\left(a_t\mid s_t\right)=\mathbb{P}\left(A_t=a_t\mid S_t=s_t\right).\label{eq:stochasticPolicy}
 \end{equation}

The ultimate goal of reinforcement learning is to find an optimal policy $\pi^\star\in\Pi$, which maps states to actions that lead to the maximization of the expected (discounted) cumulative future reward $J^\pi$
\begin{equation}
    J^\pi=\mathbb{E}_{\mathcal{P},\pi}\left[G\right],\label{eq:expectedReturn}
\end{equation}
where $G$ denotes the \emph{return}. For continuing tasks with infinite horizon, where the task never ends, $G$ is defined as the discounted cumulative reward
\begin{equation}
    G=\sum_{t=0}^\infty\gamma^tR_t,
\end{equation}
where $\gamma\in\left[0,1\right)$ is the discount factor, and is used to determine and control the importance of the future rewards. In the cases with finite horizon T (episodic tasks), the return $G$ can be modified to the undiscounted version
\begin{equation}
    G=\sum_{t=0}^TR_t.
\end{equation}
The expected return for action selection policy $\pi$ given the initial state $S_0=s$, is called the state value function $V^\pi(s)$, written as
\begin{equation}
    V^\pi\left(s\right):=\mathbb{E}_{P,\pi}\left[G\mid S_0=s\right]=\mathbb{E}_{P,\pi}\left[\sum_{t=0}^\infty\gamma^tR_t\mid S_0=s\right],\label{eq:stateValueFunction}
\end{equation}
and is related to the expected return as $J^\pi=\mathbb{E}_P\left[V^\pi\left(S\right)\right]$.

An alternative to state value function is action value function $Q^\pi\left(s,a\right)$, which considers state-action pair $\left(s,a\right)$ instead of state $s$ only. In fact, it gives the value of taking action $a$ in state $s$ under a given policy $\pi$, and is defined as
\begin{equation}
    Q^\pi\left(s,a\right)=\mathbb{E}_{P,\pi}\left[G\mid s,a\right]=\mathbb{E}_{P,\pi}\left[\sum_{t=0}^\infty\gamma^tR_t\mid s,a\right].
\end{equation}
Among all possible true value functions $V^\pi$ for different policies $\pi\in\Pi$, there exists an optimal value function $V^\star$ corresponding to an optimal policy $\pi^\star$, which is at least as large as the others defined as,
\begin{equation}
    V^\star\left(s\right):=\max_{{\pi}}V^\pi\left(s\right)\label{eq:optimalStateValue1}
\end{equation}
for all $s \in \states$. The optimal policy  $\pi^\star (s)$ for all $s \in \states$ is thus a solution to $\max_{{\pi}}V^\pi\left(s\right)$.

Similarly, an optimal action value function $Q^\star$ can be defined for taking action $a$ while being in state $s$. The optimal state value function $V^\star$ and optimal action value function $Q^\star$ are related as
\begin{equation}
    V^\star\left(s\right)=\max_{{a}}Q^\star\left(s,a\right).\label{eq:optimalStateValue2}
\end{equation}
The optimal policy $\pi^\star\left(s\right)$ can thus be written as
\begin{equation}
    \pi^\star\left(s\right)=\arg\max_{{a}}Q^\star\left(s,a\right).
\end{equation}
Here, we introduced the general notation used throughout this survey. The notation specific to each section will be introduced in its respective category. In the next section, we propose a method of categorization for exploration techniques in reinforcement learning. 

\section{Categorization of Exploratory Techniques}\label{overview}

Efficient exploration has been acknowledged as an important problem in adaptive control for quite a few decades, starting with the literature on bandit problems, eg. \citet{thompson1933likelihood}. In this survey, however, we will not discuss the bandit literature, which is vast and has been the topic of the recent book \citet{lattimore2020bandit}. Instead, we focus on sequential decision making. 
Some of the early studies that acknowledged the importance of efficient exploration in this context were delivered by \citet{mozer1989discovering,sutton1990integrated,moore1990efficient,schmidhuber1990making} and \citet{barto1991real}. A study by \citet{mozer1989discovering} showed that efficient learning of tasks modeled by a finite-state automaton is achievable using simple random exploration, provided that the number of states is small enough, while learning more complex tasks requires  a more intelligent exploration technique that accelerates the coverage of the state space. In another study, \citet{sutton1990integrated} demonstrated that the selection of sub-optimal actions (exploration) in non-stationary maze domains is essential for efficient learning, even though it may negatively affect the short-run acquisition of rewards.

Exploration techniques have been categorized in a few studies mainly based on the choice of inclusion of information in pursuing exploration. For instance, in a technical report, \citet{moore1990efficient} emphasizes the interplay between the exploration-exploitation balance and efficient learning. He categorizes exploratory moves based on the action selection method into \emph{entirely random}, \emph{local random}, and \emph{sceptical} categories. His proposed categorization states that while in the entirely-random method the agent chooses the exploratory actions totally at random, in the local-random experimentation it selects actions from the perturbed best known actions. As the very first action in the local-random method is chosen completely at random, the performance of this method is very sensitive to the quality of the first chosen action. Thus, a wrong decision at the initial step may lead to much larger learning times. The sceptical exploration method chooses actions depending on the prediction for the best known action. If it is predicted to be unsuccessful, the agent explores other actions, and selects the best known action otherwise. Although this categorization of exploration methods can explain the similarities and differences between some of the early proposed approaches, it does not provide a general foundation for classifying exploration techniques.

One of the first \emph{general} categorization of the exploration methods was introduced by \citet{thrun1992efficient}. He grouped exploration techniques into two general categories: \emph{undirected} and \emph{directed} methods. The undirected or \emph{uninformed} exploration methods do not use any sort of exploration-specific knowledge in order to perform the exploration task. These methods generally rely \emph{solely} on randomness in selecting actions. Random walk is the simplest method in this category, which is believed to be first utilized in action selection mechanisms by \citet{anderson1986learning}, \citet{munro1987dual}, \citet{mozer1989discovering}, \citet{jordan1989generic}, \citet{nguyen1990truck} and \citet{thrun1991planning}. Other examples of undirected methods consist of the exploration techniques related to Boltzmann distribution (based on utility and temperature parameter for controlling the exploration-exploitation trade-off) \citep{barto1991real, watkins1989learning, lin1992self, singh1992transfer, sutton1990integrated, lin1990self} and random action selection with a certain probability \citep{whitehead1991learning, mahadevan1992automatic, mahadevan1991scaling}. On the contrary, directed or \emph{informed} exploration methods utilize exploration-specific knowledge of the learning process to direct the agent towards exploring the environment. These exploration techniques are more efficient and beneficial compared with the undirected exploration methods in terms of complexity and cost \citep{thrun1992efficient, whitehead1991complexity}. In particular, random exploration methods may lead to an increase in the learning time as well as safety issues due to the random selection of unsafe actions repeatedly (especially in real cases, such as in robotics). Consequently, in the following years, exploration-related studies focused increasingly on ``directed'' exploration methods. The consequent diversity in these approaches necessitates the provision of a new basis for a refined categorization of these methods.

\begin{figure}[h!]
\begin{center}
\includegraphics[width=1 \textwidth]{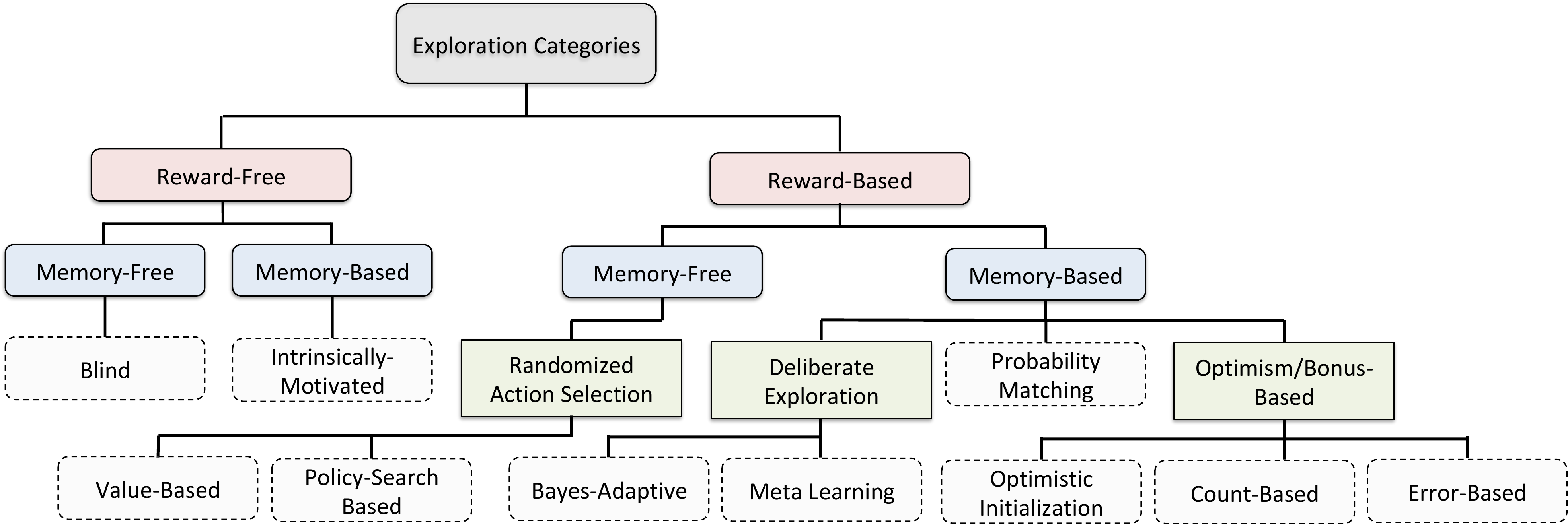}
\end{center}
\caption{Exploration Categories- The exploration methods are categorized into two main groups \emph{reward-free} and \emph{reward-based} exploration techniques, depending on their utilization of extrinsic rewards. Each group is further divided to \emph{memory-based} and \emph{memory-free} categories based on the reliance of the exploratory decisions on the agent's memory of the observed space.}
\label{fig:categories}
\end{figure}

To address the issues corresponding to the existing categorization methods, we propose a new classification approach based on the type of information the agent uses to explore the world (Figure \ref{fig:categories}). In particular, we categorize the RL exploration methods into the two general classes, namely ``Reward-Free Exploration'', where the included exploration techniques do not use the extrinsic rewards in their action selection process, and ``Reward-Based Exploration'', in which extrinsic rewards affect the choice of exploratory actions. These two classes are further divided into two groups ``Memory-Free'' and ``Memory-Based'' techniques, depending on the reliance of the exploratory movements on the agent's memory of the observed space. The categories in each class of exploration techniques are described below and detailed in the following sections.

\begin{itemize}
	\item \textbf{Reward-Free Exploration -} The general property of the exploration methods included in this category is that rewards (or value functions) do not affect the choice of actions in their action-selection criteria. In other words, actions are selected without regard to the obtained rewards or the value functions. The methods belonging in this section can either act completely blindly, which we call \emph{blind exploration}, or utilize some sort of information (other than extrinsic rewards) in the form of intrinsic rewards in order to encourage exploration. This type of exploration methods are referred to as \emph{Intrinsically-Motivated Exploration} techniques. The details regarding this category and its subcategories are provided in section \ref{sec:pure}.
	
	\item \textbf{Randomized Action Selection -} The exploration methods in this category induce exploratory behaviour via assigning action selection probabilities to the admissible actions based on the estimated value functions or rewards (\emph{Value-Based Exploration}), or the learned policies (\emph{Policy-Search Based Exploration}). The exploration methods included in the former group use the reward-based feedback in order to handle the exploration-exploitation trade-off. The list of the exploration techniques in this category as well as the detailed explanation of each method are provided in section \ref{sec:valuebased}. In the latter group, exploration methods explore the environment via performing search in the space of policies. They learn a stochastic policy, whose stochasticity helps the agent balance the trade-off between exploration and exploitation in the system. These methods explicitly represent policies, and aim to update them to maximize the expected extrinsic rewards (section \ref{sec:value-policy-search}). Note that the policy-search based methods that \emph{do not} utilize extrinsic rewards in their exploratory decision making are listed and discussed in section \ref{sec:pure}.

	\item \textbf{Optimism/Bonus-Based Exploration -} The exploration techniques in this category function based on the principle of \emph{optimism in the face of uncertainty}, where actions with uncertain values are preferred over the rest of the possible actions. In this category of exploration methods, the methods usually involve a form of \emph{bonus}, which is added to the extrinsic reward, leading to a directed search in the spaces of state-action. The methods included in this category and the details are provided in section \ref{Bonus}. The main difference between bonus-based techniques and the intrinsically-motivated exploration methods, discussed in section \ref{sec:pure}, is that the latter does \emph{not} utilize extrinsic reward for motivating the exploration of the environment. Different forms of optimism/bonus-based exploration approaches are discussed in section \ref{Bonus}, including \emph{count-based} exploration methods and \emph{prediction-error based} approaches. Finally, the clear distinction between the methods in this category and those in the \emph{Stochastic Action Selection} methods (\ref{sec:stochastic}) is that the techniques considered as \emph{Optimism/Bonus-Based} direct the agent's moves with the use of \emph{bonuses}, while the methods in the other group rely solely on the extrinsic rewards.
	
	\item \textbf{Deliberate Exploration -} This category includes exploration methods that operate based on solving the exploration-exploitation tradeoff optimally and is discussed in section \ref{sec:prior}. This category consists of \emph{Bayes-Adaptive} exploration methods that are realized with a Bayesian model-based set-up, where the \emph{posterior} distribution over models is computed and updated assuming a prior over the transition dynamics. This group also consists of \emph{Meta-Learning Based Exploration} techniques, via which the agent learns to adapt quickly using the prior given tasks.
	
	\item \textbf{Probability Matching -} 
	This category of exploration techniques 
	uses a heuristic to decide the next action based on sampling a single instance from the posterior belief over environments or value functions, and solving for that sampled environment exactly. The agent can then act according to that solution, e.g. for the duration of one episode. Each action is thus taken with the according to the probability the agent considers it to be the optimal action. This heuristic effectively directs exploration effort to promising actions.
	
\end{itemize}
In terms of the categorization proposed by \citet{thrun1992efficient}, we can classify our proposed categories into the two general groups of \emph{directed} and \emph{undirected} exploration techniques. In this regard, in the reward-free exploration category, \emph{blind} exploration methods (section \ref{sec:Random}) are undirected, while the \emph{intrinsically-motivated} exploration techniques (section \ref{sec:intrinsic}) are considered directed exploration approaches. The \emph{stochastic action selection} exploration category (section \ref{sec:valuebased}) consists of undirected techniques. The rest of the categories fall under the directed exploration category. 

In the following sections, the above mentioned groups of exploration techniques are discussed in more detail and the methods under each group are explained. In a few cases among the exploration techniques, there exist some approaches that belong to two of the proposed categories, leading to a small overlap between the groups. Whenever such situation is encountered, the respective exploration method is noted as shared between the corresponding categories.

\section{Reward-Free Exploration}\label{sec:pure}

We use the notion of \emph{reward-free} exploration in reinforcement learning to describe any method of exploration that does not incorporate extrinsic reward in their exploratory action selection criteria. This type of exploration methods was first introduced and utilized with the name \emph{pure exploration} in multi-armed bandits, a set of sequential decision-making tasks where at each time step, an agent pulls an arm and receives a random reward drawn from the reward distribution of that specific arm~\citep{bubeck2009pure}. In particular, these exploration techniques do not incorporate the rewards obtained from the environment (\emph{i.e.} extrinsic rewards) in measuring the cost of picking bandit arms. Instead, they utilize other available resources in a limited budget, such as CPU time or cost, in order to acquire knowledge. Similarly, there have been reward-free exploration methods proposed for the reinforcement learning framework, which do not rely on the extrinsic rewards the agent receives from the environment. These exploration techniques can be: 1) completely blind, where the exploratory agent selects actions in the absence of any sort of information obtained as the result of its interaction with the environment; or 2) driven by some form of intrinsic motivation and curiosity. These two forms of reward-free exploration methods are further explained, and the corresponding proposed methods are discussed in the following paragraphs and listed in Table \ref{tab:reward-free}. Note that in neither of the aforementioned cases, the agent uses extrinsic reward as a source of knowledge. Thus, bonus-based methods (\emph{i.e.} exploration methods that rely on a sort of bonus reward in \emph{addition to} extrinsic reward) do not belong in this category and are discussed in \emph{Optimism/Bonus-Based Exploration} section (section \ref{Bonus}).

\subsection{Blind Exploration}\label{sec:Random}
Exploration techniques in the \emph{blind exploration} category explore environments solely on the basis of random action selection. In other words, these agents are not guided through their exploratory path by any form of information, thus are uninformed or \emph{blind}. This category of exploration techniques is indeed the most basic type of reward-free exploration, and includes random-walk as the simplest exploration method~\citep{thrun1992efficient}. Some examples of the early uses of random walk in exploring the effect of various actions on different states were the studies performed by \citet{anderson1986learning, mozer1989discovering} and \citet{jordan1989generic}. In the random walk method, the agent chooses actions randomly regardless of the information it has obtained so far and thus, due to the uniformly random probability of selecting actions, there is a chance that the picked action takes the agent away from the goal rather than taking it closer (\emph{i.e.} exploration). On the other hand, it leads to large complexity that grows exponentially with the size of the environment~\citep{whitehead1991complexity}, which makes random-walk an inefficient exploration technique.

Another simple yet effective exploration method in this category is known as the method of $\epsilon$-greedy \citep{sutton1996generalization}, also known as \emph{max-random} or \emph{pseudo-stochastic} \citep{caironi1994training, watkins1989learning}. In the $\epsilon$-greedy approach, the parameter $\epsilon\in\left[0,1\right]$ controls the balance between exploration and exploitation. The action $a_t$ at every time step $t$ is chosen such that,
\begin{align}
a_t=\begin{cases}
a_t^\star,& \text{with probability 1-$\epsilon$} \\
random~action &\text{with probability $\epsilon$},
\end{cases}
\end{align}
where $a_t^\star$ is the greedy action taken at time $t$ with respect to the greedy policy (exploitation).

The $\epsilon$-greedy method has been found to be quite effective in different RL settings~\citep{sutton1998reinforcement}. In particular, it is efficient in the sense that it does not need to cache any data to perform exploration, and the only hyperparameter to adjust is $\epsilon$. However, despite the fact that the $\epsilon$-greedy method guarantees that at infinite time horizon every state-action pair is visited infinitely often, it stays sub-optimal in the sense that it asymptotically prevents the agent from selecting the best action \citep{vermorel2005multi}. Another problem an agent may encounter while using $\epsilon$-greedy is the lack of decisiveness in the exploration phase, which might in turn lead to getting stuck in local optima. To address this issue, a temporally extended form of $\epsilon$-greedy, called $\epsilon z$-greedy \citep{dabney2020temporally}, has been proposed, where random exploratory actions are replaced by temporally-extended sequence of actions. In particular, the $\epsilon z$-greedy agent exploits with probability $1-\epsilon$ and explores via repeating the same action for a certain number of steps $n\sim z$, where $z(n)$ is a distribution over the action-repeat duration $n$. \citet{dabney2020temporally} perform experiments in tabular as well as deep RL frameworks in tasks with discrete-action spaces, and tabular or discretized continuous state spaces.

Although $\epsilon$ is usually hand-tuned depending on the type of the problem, there are other proposed extensions of the $\epsilon$-greedy method, such as $\epsilon$-first \citep{tran2010epsilon} (where exploration is done during the first $\epsilon T$ time steps- $T$ is the total number of steps) and decreasing-$\epsilon$ in bandits \citep{caelen2007improving}, where $\epsilon$ is a decreasing function of \emph{time}, as well as the derandomization of $\epsilon$-greedy in RL tasks \citep{even2002convergence}. Another extension of the $\epsilon$-greedy approach is the \emph{Value-Difference Based Exploration} (VDBE) method (\citet{tokic2010adaptive} in Bandits and \citet{tokic2011value} in reinforcement learning), which adjusts the exploration rate $\epsilon$ based on the changes in the state-action value functions. The methods \citet{even2002convergence, tokic2010adaptive,tokic2011value}, which incorporate extrinsic rewards in their exploratory decision making, are discussed in detail in section \ref{sec:valuebased}. There are other blind exploration methods mainly proposed in the field of Robotics, for instance the spiral search technique \citep{burlington1999spiral}, which ensures visiting new locations in planar environments via expanding the search in the space with the use of logarithmic spirals. However, due to the study limit of these methods to Robotics and planar environments, we are not going to cover them here.

\subsection{Intrinsically-Motivated Exploration}\label{sec:intrinsic}
The second exploration type in the \emph{reward-free exploration} category is the \emph{intrinsically motivated exploration}, which is composed of the methods that utilize a form of intrinsic motivation in the absence of external rewards to promote exploring the unexplored parts of the environment. In contrast to blind exploration, intrinsically-motivated exploration techniques utilize some form of intrinsic information to encourage exploring the state-action spaces. The idea of employing internal incentives in exploratory tasks is borrowed from \emph{intrinsically-motivated behaviour} in humans, which has been studied and discussed extensively in education and psychology literature \citep{deci1971effects, deci1975intrinsic, amabile1976effects, benware1984quality, deci1985intrinsic, grolnick1987autonomy}. In psychology, the distinction between an extrinsically and an intrinsically motivated behaviour is made based on the types of the stimuli that ``move'' the person to perform a task \citep{ryan2000intrinsic}. While intrinsic motivation leads to an inherent satisfaction of performing a job, extrinsic motivation sets an external regulation, which encourages a person to do a task in order to obtain some separable outcome (\emph{e.g.} reward or reinforcement). Studies show that internalization of the external regulations, also known as self-regularization or self-determination \citep{deci1985intrinsic}, helps children attain higher achievements \citep{benware1984quality,grolnick1987autonomy} in terms of learning or completing complex tasks, in contrast to using extrinsic motivations in the form of rewards or reinforcement. Analogs of intrinsic motivation in human can be employed in order to promote exploration in the RL framework. Here, we review some of these studies and discuss different forms of intrinsic motivation that have been used in exploration techniques in the reinforcement learning tasks.

In the context of reinforcement learning, curiosity takes various interpretations and forms depending on the types of problems and the defined goals as well as the approaches taken toward understanding and solving the problems. In general, we can define curiosity as a way or desire to explore new situations that may help the agent with pursuing goals in the future. As agents are encountered with deceptive and/or sparse rewards in many RL set-ups, exploratory agents that rely on extrinsic rewards might end up in \emph{local optima} because of deceptive rewards or get stuck due to zero gradient in the received rewards. Thus, the techniques that intrinsically motivate the agent to explore the environment and do not rely on the extrinsic reinforcement are effective in learning of such tasks.

Many of the reward-free intrinsically-motivated exploration strategies aim at minimizing the agent's \emph{uncertainty} or \emph{error} in its predictions \citep{schmidhuber1991curious, schmidhuber1991possibility, pathak2017curiosity}. In order to evaluate the precision of the agent's predictions of the environment behavior, a model of the environment dynamics is required, such that given the current state and the chosen action, the model predicts the next state. Minimization of the resulting error in the model prediction encourages the exploration of the underlying space. There are other reward-free techniques that pursue the maximization of \emph{space coverage} (new states or state-action pairs visitation), which utilize a form of intrinsic motivation to govern the agent's exploratory behaviour \citep{hazan2018provably, amin2020locally}. More \emph{space coverage} essentially means visiting more unexplored states in a shorter amount of time and in turn, learning more about the environment. In this section, we survey and discuss some of the reward-free intrinsically-motivated exploration approaches that seek either of the above-mentioned goals. Note that the notion of \emph{intrinsic motivation} in RL tasks has been also used in combination with external rewards, which is not the subject of our discussion in this section and will be elaborated in the ``Bonus/Optimism-Based Exploration'' category (section \ref{Bonus}).

The early use of intrinsic motivation in computational framework dates back to 1976, when \citet{lenat1976artificial} used the notion of ``interestingness'' in mathematics to encourage new concepts and hypotheses. \citet{scott1989learning} introduced DIDO, a curiosity-driven learning strategy that can help the agent explore initially unknown domains in an unsupervised set-up using the notion of Shannon's uncertainty function defined as
\begin{align}
sh = -\sum_{i=1}^n\left(p_i\times\log_2(p_i)\right),
\end{align}
where $p_i$ is an estimate of the probability of the outcome $O_i$, and the summation is taken over all outcomes. Minimization of \emph{uncertainty} in their proposed formalism leads to a broader searching span, as well as more effective learning of the search space. DIDO, in fact, promotes the idea of using an experience generator that provides experiences, which are novel compared to the previous ones and are related to them at the same time. The obtained experiences help the agent search for a better representation in the space of possible representations while DIDO's representation generator is employed in finding more informative experiences. \citet{scott1989learning} performed DIDO in several discrete domains, which showed that their exploratory algorithm enables the agent to select sensible experiences and eventually leads to a good representation of the domain.

In the following years, another form of curiosity-driven exploration was introduced \citep{schmidhuber1991curious, schmidhuber1991possibility} based on the improvement in the reliability of the RL agent's predictions of the world model. In particular, \citet{schmidhuber1991possibility} proposed a model-building control system that could provide an adaptive model of the environmental dynamics. He further proposed the notion of dynamic \emph{curiosity} and \emph{boredom}, described as ``the explicit desire to improve the world model'', as a potential means of increasing the knowledge of the animat about the world in the exploration phase. In his work, \emph{curiosity} aims at minimization of the agent's ignorance and is triggered when the agent comes to the realization that it does not have enough knowledge of something. It provides a source of reinforcement for the agent and is defined as the Euclidean distance between the real and the predicted model network. A failure in the correct prediction of the environment leads to a positive reinforcement that encourages the agent to further explore the corresponding actions. Improvement in the world model predictions with time leads to less reinforcement and thus, discouragement of exploring the corresponding actions, also referred to as \emph{boredom}. 

While the idea of using curiosity was not implemented in \citet{schmidhuber1991possibility}, Schmidhuber later utilized the notion of \emph{adaptive curiosity} \citep{schmidhuber1991curious} to encourage exploration of the unpredictable parts of the environment. In particular, he proposed a curious model-building control system, where the notion of \emph{adaptive confidence} was used for modeling the reliability of a predictor's predictions, and \emph{adaptive curiosity} was utilized to reward the agent for encountering hard but learnable states and thus improve the \emph{exploration} phase by reducing the extra time spent on the non-useful or well-modelled parts of the environment. The strength of his proposed approach in comparison to its predecessors including his previous study \citep{schmidhuber1991possibility}, lies in its ability to work in \emph{uncertain non-deterministic} environments by adaptively modeling the reliability of a predictor's predictions and learning to predict \emph{cumulative} error changes in the model network. This goal can be achieved via maximization of the expectation of cumulative changes in prediction reliability. \citet{schmidhuber1991curious} tested his proposed curiosity-driven algorithm based on Watkin's Q-learning in two-dimensional \emph{discrete-state} toy environments with over 100 states and compared the results to the ones obtained using random search as the exploratory approach. Utilizing an adaptive curious agent led to a decrease of an order of magnitude in the learning time. 

Another similar yet different exploration approach was proposed by \citet{thrun1992active} around the same time, which suggested using the notion of \emph{competence map} for guiding exploration via estimating the controller's \emph{accuracy}. In particular, \citet{thrun1992active} introduced a notion of energy 
\begin{align}
E = (1-\Gamma)E_{\mbox{\scriptsize explore}} +\Gamma E_{\mbox{\scriptsize exploit}},
\end{align}
where gain parameter $0<\Gamma<1$ controls the exploration-exploitation trade-off and is a function of the change in the exploration energy $E_{\mbox{\scriptsize explore}}$ and the exploitation energy $E_{\mbox{\scriptsize exploit}}$. A competence network system is trained to estimate the upper bound of the ``model network error''; minimization of the ``expected competence'' leads to the exploration of the world. \citet{thrun1992active} employed ``competence map'' in a continuous two-dimensional robot navigation task, which revealed that their suggested exploration method can perform better compared with ``random walk'' and ``purely greedy'' approaches. 

An exploration approach was later proposed by \citet{storck1995reinforcement} as an extension of previous similar studies, such as \citet{schmidhuber1991curious, schmidhuber1991possibility, thrun1992active}. Their proposed exploration method, called \emph{Reinforcement Driven Information Acquisition} (RDIA), is devised for non-deterministic environments and utilizes the notion of \emph{information gain}, which is used as an intrinsic motivation to govern the agent’s exploratory movements. In particular, the RDIA agent models the environment via estimating the transition probability $p^\star_{ijk}(t)$ at each time step $t$ as the ratio of the number of times so far that the pair $(s_i, a_j)$ has led to the state $s_k$ over the number of times the agent has experienced $(s_i, a_j)$. The information gain is then defined as the difference between the agent’s current estimation of the transition probability $p^\star_{ijk}(t)$ and $p^\star_{ijk}(t+1)$ at time $t+1$. Information gain represents the information that the agent has acquired upon performing the respective action, which consequently leads to an increase in the estimator’s accuracy. \citet{storck1995reinforcement} assess their proposed method in simple discrete environments with certain numbers of states and actions using two different information gain measures, namely the entropy difference and the Kullback-Leibler (KL) distance, between the probability distributions, and show that the results obtained by RDIA surpass those of simple random search.

Information gain as intrinsic motivation has been used in other exploration strategies such as the studies performed by \citet{little2013learning} and \citet{mobin2014information}. In particular, the exploration method proposed by \citet{little2013learning} takes a Bayesian approach, where the agent builds an internal model of the environment, and upon taking an action and observing the next state, calculates the KL-divergence of its current internal model from the one it had predicted prior to taking the action. The resulting unweighted sum of the KL-divergences yields the \emph{missing information} $I_M$, which is utilized as a measure of inaccuracy in the agent’s internal model. The agent subsequently takes actions that maximize the expected (predicted) information gain (PIG), defined as the expected decrease in the missing information $I_M$ between the internal models.  Another similar exploration method \citep{mobin2014information} extends the application of PIG \citep{little2013learning} to perform in environments with \emph{unbounded} discrete state spaces. In particular, \citet{mobin2014information} utilize the Chinese Restaurant Process (CRP)\citep{aldous1985exchangeability} to find the probability of revisiting a state or discovering a new one, and use the obtained results to calculate the agent’s internal model. The subsequent steps are similar to those presented and discussed by \citet{little2013learning}. Another study by \citet{shyam2019model} introduces the Bayesian Model-based Active eXploration (MAX) method, which utilizes the \emph{novelty of transitions} as a learning signal, and is applicable in discrete and continuous environments. In particular, MAX agent calculates the Jenson-Shannon divergence and the Jensen-R\'enyi divergence \citep{renyi1961measures} of the predicted space of distributions from the resulting one in discrete and continuous environments, respectively. Maximization of the resulting novelty measure governs the agent’s exploratory behaviour. The evaluation of MAX performance in several discrete and continuous tasks presents promising results compared with those obtained from MAX counterparts and other baselines.

\begin{table}
    \centering
    \begin{tabular}{p{4.9cm}|p{2.5cm} |p{7cm}}
        Approach & Intrinsic  & Remarks \\
         & Motivation & \\
        \hline
        \cite{anderson1986learning}& None (Blind) & Early use of \emph{random walk}
        \\
        \cite{mozer1989discovering}& None (Blind) & Early use of \emph{random walk}
        \\
	\cite{jordan1989generic}& None (Blind) & Early use of \emph{random walk}
	\\
	\cite{sutton1996generalization}& None (Blind) & $\epsilon$-greedy
	\\
	\cite{caironi1994training}& None (Blind) & max-random ($\epsilon$-greedy)
	\\
	\cite{dabney2020temporally}& None (Blind) & $\epsilon z$-greedy (Temporally-extended actions)
	\\
	\cite{burlington1999spiral}& None (Blind) & Spiral search (For planar environments only)
	\\
	\cite{schmidhuber1991possibility}& Uncertainty & Adaptive model of the environment dynamics
	\\
	\cite{scott1989learning}& Uncertainty & Minimization of Shannon's uncertainty function
	\\
	\cite{schmidhuber1991curious}& Uncertainty & Prediction of cumulative error changes
	\\
	\cite{thrun1992active}& Uncertainty & Competence map
	\\
	\cite{storck1995reinforcement}& Uncertainty & Maximization of information gain
	\\
	\cite{little2013learning}& Uncertainty & Maximization of expected information gain
	\\
	\cite{mobin2014information}& Uncertainty & Maximization of expected information gain
	\\
	\cite{shyam2019model}& Uncertainty & Uses an ensemble of forward dynamics models
	\\
	\cite{pathak2017curiosity}& Uncertainty & Minimization of predicted error in feature representation
	\\
	\cite{hazan2018provably}& Space coverage & Maximization of entropy of the distribution over the visited states
	\\
	\cite{amin2020locally}& Space coverage & Generation of correlated trajectories in the state and action spaces
	\\
	\cite{forestier2017intrinsically}& Self-generated goals & Coverage maximization of the space of goals
	\\
	\cite{colas2018gep}& Self-generated goals & Combines \citet{forestier2017intrinsically} with DDPG
	\\
	\cite{machado2017laplacian}& Space coverage & Maximization of eigenpurposes
	\\
	\cite{c.2018eigenoption}& Space coverage & Extension of \citet{machado2017laplacian} to stochastic environments
	\\
	\cite{jinnai2019discovering}& Space coverage & Minimization of cover time
	\\
	\cite{jinnai2020exploration}& Space coverage & Extension of \citet{jinnai2019discovering} to large or continuous state spaces
	\\
	\cite{hong2018diversity}& Space coverage & Encouraging new policies using a distance measure between the policies

    \end{tabular}
    \caption{Examples of some reward-free exploration approaches.}
    \label{tab:reward-free}
\end{table}

Another curiosity-driven approach is ``Intrinsic Curiosity Module'' (ICM) \citep{pathak2017curiosity}, where curiosity is defined as ``the error in an agent's ability to predict the consequence of its own actions''. In their set-up, the agent interacts with high-dimensional continuous state spaces (images in this case). The authors show that ICM helps the agent to learn and improve its exploration policy in the presence of sparse extrinsic rewards as well as in the absence of any sort of environmental rewards. Moreover, they show that the curious agent can apply its gained knowledge and skills in new scenarios and still achieve improved results. The main idea behind ICM is that instead of targeting \emph{learnable} states and rewarding the agent for detecting them \citep{schmidhuber1991possibility, schmidhuber1991curious}, ICM \citep{pathak2017curiosity} focuses only on a \emph{feature representation} that reflects the parts of the environment that either affect the agent or get affected by the agent's choice of actions. Intuitively, by focusing on the influential feature space instead of the state space, ICM is able to avoid the unpredictable or unlearnable parts of the environment.

Note that in some of the fore-mentioned proposed algorithms \citep{schmidhuber1991possibility, schmidhuber1991curious, pathak2017curiosity}, while the external reinforcement is not a necessary component, the external reward, if exists, can be added to the curious reinforcement. Thus, these studies are included in the list of ``Bonus/Optimism-Based Exploration'' category as well (section \ref{Bonus}). 

Some of the intrinsically-motivated exploration techniques utilize the analogy between the dynamical and physical systems, and thus propose the notion of \emph{entropy maximization} to encourage exploration of the search space. In this regard, an early utilization of entropy in intelligent adaptation of search control was in the context of \emph{search effort allocation} problem in genetic search procedures \citep{rosca1995entropy}, where entropy was used as a measure of diversity. Around the same time, \citet{wyatt1998exploration} defined a notion of entropy for the case of bandits as a measure of the uncertainty regarding the identity of the optimal action. In his proposed exploration algorithm, the agent selects the more informative action, which is the one that on expectation will lead to a larger entropy reduction. In the field of reinforcement learning (which is the main focus of the current survey), there are several studies that utilize notion of entropy in their proposed exploration techniques \citep{achbany2006optimal, lee2018sparse, yin2002maximum, hazan2018provably}. In this section, however, we discuss the method proposed by \citet{hazan2018provably} as it is the only \emph{reward-free} technique among the studies that utilize the notion of entropy in guiding exploration. \citet{hazan2018provably} introduce an exploration approach, which targets environments that do not provide the agent with extrinsic reward. In their proposed method, the intrinsic objective is to maximize the entropy of the distribution over the visited states. They introduce an algorithm, which optimizes objectives that are only functions of the state-visitation frequencies. In particular, it generates and optimizes a sequence of intrinsic reward signals, which consequently leads to the entropy maximization of the distribution that the policy induces over the visited states. The reward signals form a concave reward functional $R(d_\pi)$, which is a function of the entropy of the induced state distribution $d_\pi$ given policy $\pi$. The obtained optimal policy is referred to as \emph{maximum-entropy} (MaxEnt) exploration policy, which is defined as $\pi^\star \in \argmax_{\pi} R(d_\pi)$.

Another intrinsically-motivated exploration approach that encourages space coverage is the method of PolyRL \citep{amin2020locally}, which is designed for tasks with continuous state and action spaces and sparse reward structures. PolyRL is inspired by the statistical models used in the field of polymer physics to explain the behaviour of simplified polymer models. In particular, PolyRL exploration policy selects orientationally correlated actions in the action space and induces persistent trajectories of visited states (\emph{locally self-avoiding} walks) in the state space using a measure of spread known as the \emph{radius of gyration squared},
\begin{align}
    U_g^2(\tau_\states) &:= \frac{1}{T_e-1} \sum_{s \in \tau_\states} d^2(s,\Bar{\tau}_\states).\label{eq:radius_gyration}
\end{align}
In equation \ref{eq:radius_gyration}, $T_e$ denotes the number of exploratory steps taken so far in the current exploratory trajectory, $\tau_\states$ is the trajectory of the visited states, and $d(s,\bar{\tau}_\states)$ is a measure of distance between a visited state $s$ and the empirical mean of all visited states $\Bar{\tau}_\states$. At each time step, the exploratory agent computes  $U_g^2(\tau_\states)$, and subsequently compares it with the value obtained from the previous step. In addition, it calculates the high-probability confidence bounds on the radius of gyration squared, within which the stiffness of the trajectory is maintained. If the change in $U_g^2(\tau_\states)$ is within the confidence interval, the agent continues to explore, otherwise it selects the subsequent action using the target policy. \citet{amin2020locally} assess the performance of PolyRL in 2D continuous navigation tasks as well as several high-dimensional sparse MuJoCo tasks and show improved results compared with those obtained from several other exploration techniques. \\

\textbf{Policy-Search Based Exploration without Extrinsic Reward -}
\label{sec:pure-policy-seach}\emph{Policy-search} methods search in the parameter space $\theta$ for the appropriate parameterized policy $\pi_\theta$. As policy-search methods typically do not learn value functions, the choice of a proper parameter $\theta$ is essential for ensuring an efficient, stable and robust learning. This calls for an efficient exploration strategy in order to provide the policy evaluation step in the policy search methods with new trajectories and thus new information, which is subsequently used for policy update \citep{deisenroth2013survey}. The exploration approaches performed in policy search methods use stochastic policies, and they can either employ rewards obtained from the environment to guide the exploratory trajectories or function in a completely reward-free manner. In the current section, we review the policy-search methods that do \emph{not} incorporate extrinsic rewards in their exploratory decision making. We provide a more thorough introduction to policy-search methods in section \ref{sec:value-policy-search}, where we discuss the policy-search approaches that utilize extrinsic rewards.

One of the approaches to solving problems autonomously is breaking the problem/goal into smaller sub-problems/sub-goals \citep{forestier2017intrinsically}. This idea is inspired from the way children tend to select their objectives such that they are not too easy or too hard for them to handle. These intermediate learned goals facilitate learning more complex goals, which ultimately lead to building up more skills required to achieve bigger goals. Based on this intuition, \citet{forestier2017intrinsically} propose a curiosity-driven exploration algorithm called ``Intrinsically Motivated Goal Exploration Process'' (IMGEP). The IMGEP approach structure relies on assuming that the agent is capable of choosing goal $p$ from the space of RL problem and is able to calculate the corresponding reward $r$ using the reward function $R\left(p,c,\theta,o_\tau\right)$ given the parameterized policy $\pi_\theta$, context $c$ (which gives the current state of the environment) and the observed outcome $o_\tau$ in the trajectory $\tau=\{ s_{t_0},a_{t_0}, s_{t_1}, a_{t_1},\dots,s_{t_{\mbox{\scriptsize end}}}, a_{t_{\mbox{\scriptsize end}}}\}$. The reward function $R\left(p,c,\theta,o_\tau\right)$ is thus non-Markovian and can be calculated at any time during or after performing the tasks. Using the computed rewards, the agent samples the \emph{interesting} goal $p$, which is a self-generated goal that leads to faster learning progress. In the exploration phase, the agent uses the meta policy $\Pi_\epsilon\left(\theta|p,c\right)$ to find the parameter $\theta$ for goal $p$, which is subsequently utilized in a goal-parameterized policy search process. The obtained outcome is then used in computing the intrinsic reward $r$, which in turn provides useful information regarding the interestingness of the samples goal $p$. The goal sampling strategy and the meta-policy are subsequently updated. The performance of IMGEP in the case of a real humanoid robot shows that the IMGEP robot can effectively explore high-dimensional spaces through discovering skills with increasing complexity. 

One of the exploration methods proposed based on the ``Goal Exploration Processes'' (GEPs) \citep{forestier2017intrinsically} is the ``Goal Exploration Process- Policy Gradient'' (GEP-PG) \citep{colas2018gep}, which combines the intrinsically-motivated exploration processes GEPs with the deep reinforcement learning method DDPG in order to improve exploration in continuous state-action spaces and learn the tasks. \citet{colas2018gep} perform GEP-PG in the low-dimensional ``Continuous Mountain Car'' and the higher-dimensional ``Half-Cheetah'' tasks. GEP-PG is tested in the fore-mentioned problems with different variants of DDPG. The authors show that specifically in the Half-Cheetah task, the performance, variability and sample efficiency of their proposed method surpasses those of DDPG.

Another policy-search based exploration strategy proposed by \citet{machado2017laplacian} utilizes the notion of \emph{proto-value functions} (PVFs) to discover \emph{options} that lead the agent towards efficient exploration of the state space. PVFs, first introduced by \citet{mahadevan2005proto}, are the basis functions used for approximating value functions through incorporating topological properties of the state space. In particular, using the MDP's transition matrix, a diffusion model is generated, whose diagonalized form subsequently gives rise to PVFs. The diffusion model provides the diffusion information flow in the environment. This feature allows the PVFs to provide useful information regarding the geometry of the environment, including the \emph{bottlenecks}. \citet{machado2017laplacian} define an intrinsic reward function (a.k.a. \emph{eigenpurpose}) as,
\begin{align}
r^{\textbf e}_{\mbox{\scriptsize in}}\left(s,s^\prime\right) = {\textbf e}^{\intercal}\left(\phi\left(s^\prime\right)-\phi\left(s\right)\right),\label{eq:eigenpurpose}
\end{align}
where ${\textbf e}\in\mathbb{R}^{|\mathcal{S}|}$ is the proto-value function and $\phi(s)$ is the feature representation of state $s$, which can be replaced by the state $s$ itself in tabular cases. Machado \emph{et al.} utilize eigenpurpose $r_i^{\textbf e}$ to discover options (\emph{a.k.a.} eigenoptions) and their corresponding eigenbehaviors. They subsequently use policy iteration to solve the problem for an optimal policy. 

Although the exploration technique introduced by \citet{machado2017laplacian} is applicable to discrete domains only, one of its major advantages is that it provides a dense intrinsic reward function, which facilitates exploration tasks with sparse-extrinsic-reward structures. Moreover, since it is equipped with options, their proposed method can cover a relatively larger span in the state space compared with that of a simple random walk. Finally, the authors show that their method with options is effective for exploration tasks with the goal of maximizing the cumulative rewards. Later work by \citet{c.2018eigenoption} proposed an improved version of \emph{eigenoption discovery}, extending it to stochastic environments with non-tabular states. In particular, using the notion of successor representation \cite{dayan1993improving}, in their proposed method the agent learns the non-linear representation of the states which in turn gives the diffusive information flow (DIF) model, and subsequently, the eigenpurposes and eigenoptions are obtained. 

In the recent years, \citet{jinnai2019discovering} introduced the method of \emph{covering options}, where options are generated with the goal of minimizing cover time. Their proposed method encourages the agent to visit less-explored regions of the state space by generating options for those parts, without using the information obtained from extrinsic rewards. Their empirical evaluation in discrete sparse-reward domains present reduced learning time in comparison with that of some of their predecessors. The method of \emph{deep covering options} \citep{jinnai2020exploration} extends covering options to large or continuous state spaces while minimizing the agent's expected cover time in the state space. The authors have successfully shown the behaviour of deep covering options in challenging sparse-reward tasks, including Pinball, as well as some MuJoCo and Atari domains.

In a study by \citet{hong2018diversity}, the authors apply a diversity-driven method to off- and on-policy DRL algorithms and improve their performances in large state spaces with sparse or deceptive rewards via encouraging the agent to try new policies. In particular, the agent uses a distance measure to evaluate the novelty of $\pi$ in comparison to the prior ones and subsequently modifies the loss function,
\begin{align}
L_D = L - \E_{\pi^{\prime}\in\Pi^{\prime}}\left[\alpha D\left(\pi,~\pi^{\prime}\right)\right], \label{eq:div}
\end{align}
where $L$ denotes the loss function of the deep RL algorithm, $\alpha$ is a scaling factor, and $D$ is a distance measure between the current policy $\pi$ and the policy $\pi^{\prime}$ sampled from a set of most recent policies $\Pi^{\prime}$. Equation \ref{eq:div}, encourages the agent to try new policies and explore unvisited states without relying on the extrinsic rewards received from the environment. The agent will consequently overcome the problem of getting stuck in local optima due to the presence of deceptive rewards or failing to learn tasks with sparse rewards or large state spaces. The authors apply their proposed exploration method in 2D grid worlds with deceptive or sparse rewards, Atari 2600 as well as MuJoCo, and show that it enhances the performance of DRL algorithms through a more effective exploration strategy.

\section{Randomized Action Selection}
\label{sec:stochastic}
So far, we have introduced and discussed exploration methods that do not acquire information in the form of extrinsic rewards in the process of exploratory action selection. In this section and the rest of the survey, we focus on the exploration techniques that make decisions using extrinsic rewards with or without other forms of information obtained from the learned process. In this section, we specifically target the exploration methods that assign action selection probabilities to the admissible actions based on value functions/rewards or the learned policies. The two groups of exploration methods are introduced in sections \ref{sec:valuebased} and \ref{sec:value-policy-search}, respectively.

\subsection{Value-Based Methods}\label{sec:valuebased}
A simple way to deal with the exploration-exploitation trade-off is to induce exploratory behaviour via assigning action selection probabilities to the admissible actions based on the estimated value functions. In the early phase of learning, the agent should be able to try different actions in each state. Later in the intermediate learning phase, if the agent's target policy takes the control of the action selection process, it may lead to a partial visitation of the state space, and thus a sub-optimal policy and value function. To tackle this issue, there are exploration approaches that select the stochastic actions based on the feedback they receive, in the form of value function or rewards, from the environment. Using these feedback, they balance exploration and exploitation via deciding between acquiring more knowledge from the environment and maximizing the obtained rewards. In this section, some of the exploration methods that perform action selection according to the above-mentioned criteria are discussed below. Examples of some value-based randomized action selection exploration methods are provided in Table \ref{tab:valueBased}.

 One of the most important exploration methods in this category is the \emph{Softmax} action selection method \citep{bridle1990training}. In this method, the greedy action at the current state is selected with the highest probability, while the other actions are given the probability of being selected according to their estimated values. The most commonly used formalism for performing Softmax action selection is the \emph{Boltzmann} distribution. The early use of Boltzmann distribution in exploration was by \citet{watkins1989learning, lin1992self} and \citet{ barto1991real}. In the Boltzmann exploration approach, the value function $Q_t(s,a)$ for the current state $S_t=s$ and action $a_t=a$ assigns the probability of selecting action $a$ as, 
\begin{align}
\pi_t\left(a|s\right)=\frac{e^{Q_t\left(s,a\right)/T_m}}{\sum_{i=1}^{N}e^{Q_t\left(s,a_i\right)/T_m}},\label{eq:Softmax}
\end{align}
where $T_m>0$ is called the \emph{temperature} and controls how frequent the agent will choose random actions as opposed to the best actions $a_t^\star = \argmax_aQ_t(s,a)$. If $T_m$ decreases, the probability of generating the action with the highest expected reward $a_t^\star$ increases, leading to a decrease in the probability of selecting other actions and hence a lower probability of exploring the environment. In the limit of \emph{zero temperature} $T_m\rightarrow0$, the agent uses the target policy to select greedy actions in order to maximize the obtained rewards. At very large temperatures $T_m\rightarrow\infty$, all of the actions have almost the same probability and the action selection process approaches random walk. Setting the value for the temperature $T$ is not straightforward, but it is generally reduced during the experiments, leading to more exploitation over time.

\begin{table}
    \centering
    \begin{tabular}{p{5cm}|p{7cm}}
        Approach &  Remarks \\
        \hline
        \cite{bridle1990training}& Softmax
        \\
        \cite{watkins1989learning}& Early use of Boltzmann distribution
        \\
        \cite{lin1992self}& Early use of Boltzmann distribution
        \\
        \cite{barto1991real}& Early use of Boltzmann distribution
        \\
        \cite{vamplew2017softmax}& Vectorization of rewards in multi-objective tasks
        \\
        \cite{wiering1999explorations}& Max-Boltzmann ($\epsilon$-greedy $+$ Boltzmann)
        \\
        \cite{tokic2010adaptive}& Value-Difference Based Exploration (VDBE)
        \\
        \cite{tokic2011value}& Extension of VDBE (VDBE $+$ Max-Boltzmann)
        \\
        \cite{tijsma2016comparing}& Controls and increases the probability of greedy action selection with time
        \\
        \cite{even2002convergence}& Derandomization of $\epsilon$-greedy

    \end{tabular}
    \caption{Examples of value-based randomized action selection exploration methods.}
    \label{tab:valueBased}
\end{table}

One of the extended applications of the \emph{Softmax} exploration method is in the multiobjective reinforcement learning tasks \citep{vamplew2017softmax}, where scalar rewards are replaced with vector-valued rewards. Each element in the vectors represent the reward corresponding to an objective. The action-value function $Q(s,a)$ is consequently presented in the form of vector $\bar{Q}(s,a)$. In order to use the vector representation of the value functions in the Boltzmann formalism, the vectors must be mapped to scalar values using a scalarization function $f(\bar{Q}(s,a))$ \citep{liu2015multiobjective}, which can have either linear \citep{vamplew2017softmax, guo2009reinforcement, perez2009responsive} or non-linear \citep{gabor1998multi, van2013hypervolume, van2013scalarized} representations. The Boltzmann formalism is consequently written as,
\begin{align}
\pi\left(a|s\right)=\frac{e^{f\left(\bar{Q}_t\left(s,a\right)\right)/T}}{\sum_{i=1}^{N}e^{f\left(\bar{Q}_t\left(s,a_i\right)\right)/T}}.
\end{align}

Another extension of Softmax exploration is the \emph{Max-Boltzmann} rule \citep{wiering1999explorations}, which is a combination of the $\epsilon$-greedy or Max-random approach (explained in detail in section \ref{sec:Random}) and the Boltzmann exploration method. In the Max-Boltzmann method, similar to the $\epsilon$-greedy approach, the action that gives the maximum Q-value is chosen with the probability $1-\epsilon$. With the probability $\epsilon$, the Boltzmann distribution (equation \ref{eq:Softmax}) is used for action selection. A drawback for the Max-Boltzmann exploration method compared to the two approaches, Max-Random and Boltzmann, is the need for tuning two hyperparameters $T$ and $\epsilon$ instead of one. However, the Max-Boltzmann method has been shown to reduce the weight of exploration in comparison with exploitation, and thus avoid over-exploration.

Another exploration technique in this category is the incremental Q-learning algorithm \citep{even2002convergence}, which is an extension of the $\epsilon$-greedy method (discussed in section \ref{sec:pure}). In particular, in their proposed method, \citet{even2002convergence} derandomize the $\epsilon$-greedy method by adding a \emph{promotion} term to the estimated Q-value for each state-action pair. If action $a$ is not taken in state $s$ at time $t$, the promotion term of that specific state-action pair at time $t+1$ is increased by a value called \emph{promotion function}, and zeroed otherwise. The promotion function plays the role of a decreasing $\epsilon$ in the $\epsilon$-greedy approach, and decreases with the number of times action $a^\prime\neq a$ has been taken in state $s$. Consequently, the values of actions that have not been taken are promoted and the fraction of time the sub-optimal actions are chosen decreases with time and vanishes in the limit.

Another extension of the $\epsilon$-greedy method is the Value-Difference Based Exploration (VDBE) \citep{tokic2010adaptive}, where the $\epsilon$ parameter in the $\epsilon$-greedy approach is replaced with a state-dependent exploration probability $\epsilon_t(s)$ instead of being hand-tuned. In VDBE, the initialization of the exploration probability $\epsilon_t(s)$  is done with $\epsilon_o(s)=1$ for all states. At each time-step, the TD-error is computed, which serves as a measure for the agent's uncertainty. A larger TD-error for a state corresponds to a larger uncertainty, which consequently triggers a higher chance of exploration by assigning a larger value to the exploration rate $\epsilon_t(s)$ for that specific state. 

Although \citet{tokic2010adaptive} assesses VDBE in multi-armed bandit problems, he argues that the method of VDBE is also applicable in multi-state MDPs. In another study, \citet{tokic2011value} introduce an extension of VDBE, namely \emph{VDBE-Softmax}, for solving reinforcement learning problems with multiple states. The VDBE-Softmax method combines VDBE (introduced by \citet{tokic2010adaptive}) with Max-Boltzmann exploration (proposed by \citet{wiering1999explorations}) in order to overcome the shortcomings of the two former approaches. A disadvantage of VDBE, as stated by \citet{tokic2011value}, is that it does not discriminate between exploratory actions, which leads to equal probability of selecting the actions that yield high and low Q-values. Another disadvantage of VDBE is its divergence in the cases where oscillations exist in the value functions caused by stochastic rewards or function approximators. In the VDBE-Softmax approach, with probability $\epsilon_t(s)$, the agent selects the exploratory actions using equation \ref{eq:Softmax}, and chooses the argmax of Q-value (greedy action) with the probability $1-\epsilon_t(s)$. The authors show that in general, their proposed variation of VDBE performs better than the exploration methods $\epsilon$-greedy, Softmax and pure VDBE in the environments with deterministic rewards (cliff-walking problem and bandit-world tasks) as well as the ones with stochastic rewards (bandit-world tasks).

Similar to the method of VDBE \citep{tokic2010adaptive}, there exist other exploration strategies, which were originally proposed for and applied in the bandit problems, but were later performed in multi-state MDPs as well. One of these approaches is called \emph{Pursuit} \citep{thathachar1984class}, which maintains the probability of selecting greedy action as well as the value of different actions at each time step. As described in \citet{sutton1998reinforcement}, in $k$-armed bandit problems, Pursuit algorithms initialize the probability of choosing an arm with $p_{t=0}(a)=1/k$ for all actions $a=1,\dots,k$. At each time step $t$, the probability of selecting action $a$ is calculated as
\begin{align}
p_{t+1}(a)=\begin{cases}
p_t(a)+\alpha\left(1-p_t(a)\right) & \text{if $a=\argmax_iQ_t(i)$}\\
p_t(a)+\alpha\left(0-p_t(a)\right) & \text{Otherwise}\label{eq:pursuit-bandit}
\end{cases}
\end{align}
where $\alpha\in(0,1)$ is the learning rate. According to equation \ref{eq:pursuit-bandit}, the probability of selecting the greedy action increases with time, leading to fewer exploratory moves and more exploitation. The equivalence of equation \ref{eq:pursuit-bandit} for the case of multi-state MDPs is given as follows \citep{tijsma2016comparing}
\begin{align}
\pi_{t+1}(s_t,a_{t})=\begin{cases}
\pi_t(s_t,a_{t})+\alpha\left(1-\pi_t(s_t,a_t)\right) & \text{if $a_t=a^\star_t$}\\
\pi_t(s_t,a_t)+\alpha\left(0-\pi_t(s_t,a_t)\right) & \text{Otherwise}\label{eq:pursuit-RL}
\end{cases}
\end{align}
where $a^\star_t=\argmax_aQ(s_t,a)$. \citet{tijsma2016comparing} perform pursuit as well as other exploration techniques including Softmax and $\epsilon$-greedy in random discrete stochastic mazes with one optimal goal and two sub-optimal goal states. Their results show that Softmax outperforms the other exploration strategies and that the $\epsilon$-greedy method has the worst performance of all. 
\subsection{Policy-Search Based Methods}\label{sec:value-policy-search}

After having discussed methods that implicitly represent a policy using value function, we turn our attention to policy search methods.
Policy search methods explicitly represent a policy, instead of, or in addition to, a value function. In the latter case, these methods are more specifically referred to as actor-critic methods. Most policy search methods learn a stochastic policy. The stochasticity in the policy is usually also the main driver of exploration in such methods.  The different ways in which such perturbations can be applied will be the focus of the next several sections\footnote{A more general overview of the properties of policy search methods is given in~\cite{deisenroth2013survey}.}: after this short introduction, Section~\ref{ssec:perturbed} will introduce the organizational principle for the section, with Sections~\ref{ssec:actionspace} and~\ref{ssec:parameterspace} explaining the individual methods in detail. 

Many policy search methods belong to the policy gradient family. These methods aim to update the policy in the direction of the gradient of the expected return $\nabla J^\pi$. A basic way to do so is by calculating a finite-difference approximation of the gradient. In this approach, rollouts are performed for one or more perturbations of the original parameter vector, which are then used to estimate the gradient. When the system is non-deterministic, these estimates are extremely noisy, although better estimates can be obtained in simulation when the stochasticity of the environment is controlled by fixing the sequence of pseudo-random numbers~\citep{ng2000pegasus}. 

More sophisticated approaches are based on the log-ratio policy gradient \citep{williams1992simple}. The log-ratio policy gradient relies on stochastic policies, and exploits the knowledge of the policy's score function (gradient of the log-likelihood). Stochastic policies for continuous actions based on the Gaussian distribution~\citep{williams1992simple} are still frequently used~\citep{deisenroth2013survey}. For discrete action spaces, a Gibbs distributions with a learned energy function~\citep{sutton2000policy} can be used instead.

The initialization of the exploration policy can be freely chosen. In some policy architectures, the amount of exploration is fixed to some constant or decreased according to a set schedule. In other architectures, the amount of exploration is controlled by learned parameters, possibly separate from other  parameters (such as those controlling the `mean action' for any given state).  Policy search methods typically maximize the expected return, and thus probability mass tends to slowly be shifted towards a more greedy policy (usually resulting in a decreasing amount of exploration). These and more advanced strategies will be discussed in more detail in Sec. \ref{ssec:policydist}. 

The discussed approaches differ in an important aspect: while in finite-difference methods \citep{ng2000pegasus} the \emph{parameters} of the policy are perturbed, the method proposed by \citet{williams1992simple} selects the \emph{actions} stochastically. Where perturbations are applied at the level of parameters, they often affect an entire episode (\emph{episode-based} perturbations). In contrast, classically action-space perturbations are often only applied for a single time step (\emph{independent} perturbations). 

In this section, we will focus on research in the area of policy search methods that  introduce new exploration strategies or that explicitly evaluate the effects of different exploration strategies. 
We will focus on such policy search methods that are trained on a single task and where the policy has its own representation. Policies that are defined only in terms of value function are covered in Section~\ref{sec:valuebased}. Policies explicitly optimizing over a distribution of tasks are covered with Bayesian and meta-learning approaches in Section~\ref{sec:prior}. An overview of the methods we will cover in this section is given in Table~\ref{tab:policysearch}. The table groups the method by type and coherence of perturbation that, like \citet{deisenroth2013survey}, we consider to be key characteristics of exploration strategies in policy search. The following subsection will give a more detailed explanation of these characteristics.

\begin{table}
    \centering
    \begin{tabular}{c|p{0.2\textwidth} |p{0.17\textwidth}|p{0.17\textwidth}}
        Approach & Perturbed space & Temporal coherence\textsuperscript{\textasteriskcentered} & Remarks \\
        \hline
        \cite{barto1983neuronlike}& Action-space & Independent & -
        \\
        \cite{gullapalli1990stochastic}& Action-space & Independent & -
        \\
        \cite{williams1992simple} & Action-space & Independent & - \\
        \cite{morimoto2001acquisition} & Action-space & Correlated & Multi-modal (hierarchy) \\
        \cite{nachum2019does} & Action-space & Correlated & - \\
        \cite{wawrzynski2015control} & Action-space & Correlated & - \\
        \cite{lillicrap2016continuous} & Action-space & Correlated & -\\
        \cite{haarnoja2017reinforcement} & Action-space & Independent & Multi-modal\\ 
        \cite{xu2018learning} & Action-space & Independent & - \\
        \hline
        \cite{kohl2004policy} & Parameter-space & Episode-based & - \\
        \cite{sehnke2010parameter} & Parameter-space & Episode-based & - \\
        \cite{ruckstiess2010exploring} & Parameter-space & Episode-based & -\\
        " & Action-space & Episode-based & - \\
        \cite{theodorou2010generalized} & Parameter-space & Episode-based & - \\
        \cite{stulp2012path}  & Parameter-space & Episode-based & Correlated \mbox{parameters}\\
        \cite{salimans2017evolution} & Parameter-space & Episode-based & -\\
        \cite{conti2018improving} & Parameter-space & Episode-based & - \\
        \cite{hoof2017generalized} & Parameter-space\textsuperscript{\textdagger} & Correlated\textsuperscript{\textdagger} & -  \\
        \cite{plappert2018parameter} & Parameter-space\textsuperscript{\textdagger} & Episode-based\textsuperscript{\textdagger} & - \\
        \cite{fortunato2018noisy} & Parameter-space\textsuperscript{\textdagger} & Episode-based\textsuperscript{\textdagger} & -\\
        \citet{mahajan2019maven} & Parameter-space & Episode-based & Multi-agent \\
    \end{tabular}
    \caption{Different exploration approaches proposed in the context of policy search algorithms. 
    The first section of the table lists methods that mainly perturb the policy in the action space, these methods will be discussed in Sec.~\ref{ssec:actionspace}. The second section lists methods that mainly perturb the policy in the parameter space, that will be discussed in Sec.~\ref{ssec:parameterspace}. Within these two broad categories, papers are ordered roughly chronologically, although papers within a similar line of work are kept together. Multiple entries for the same paper refer to different variants. \\
    \textsuperscript{\textasteriskcentered} Denotes whether perturbations are applied \emph{independently} at each timestep, don't change at all throughout an \emph{episode}, or have an intermediate \emph{correlation} structure. Details in Sec.~\ref{ssec:perturbed}.\\ 
    \textsuperscript{\textdagger} These methods have additional step-based action-space noise for numeric reasons or to ensure a differentiable objective.  }
    \label{tab:policysearch}
\end{table}

\subsubsection{Perturbed space and coherence}
\label{ssec:perturbed}
In policy based methods, exploratory behavior is usually obtained by applying random perturbations. One of the main characteristics that differentiate exploration methods is where those perturbations are applied. Within policy gradient techniques, there are two main candidates: either the actions or the parameters are perturbed (although we will discuss some approaches beyond these two shortly). These possibilities reflect two views on the learning problem. On the one hand, the actions that are executed on the system are what actually affects the reward and the next state, no matter which parameter vector generated the actions. From this perspective, it is more straightforward to start with finding good actions, and subsequently find a parametric policy that can generate them. From another perspective, parametrized policies have limited capacity, and the resulting inductive bias might mean that the true optimal policy is excluded from the set of representable policies. We are then looking for parameters with which the overall policy behavior is best across all states. Also, if the structure of the policy is chosen to  reflect prior information about the structure of the solution (e.g. policies linear in hand-picked features or policies with a hierarchical structure), perturbing the policy parameters ensure that explorative behavior follows the same structure. 

The space in which explorative behaviors are applied is usually closely linked to the coherence of behavior. Coherence here refers to the question of whether (and how closely) perturbations in subsequent time steps depend on one another. Exploration with a low coherence (e.g., perturbations chosen independently at every time step) has the advantage that many different strategies might be tried within a single episode. On the other hand, exploration with a high coherence (e.g., perturbing the policy at the beginning of each episode only) has the advantage that the long-term effect of following a certain policy can be evaluated~\citep{sehnke2010parameter}. Whereas independent perturbations can result in inefficient random walk behavior,  following a perturbed policy consistently could result in reaching a greater variety of states~\citep{machado2017laplacian}. Intermediate strategies between the extremes of completely identical perturbation across an entire episode and completely independent perturbation per time step are also possible~\citep{morimoto2001acquisition,wawrzynski2015control}, and can be used to compromise between the advantages of the more extreme strategies. 

Most exploration approaches which perturb the policy in action space have focused on independent perturbations in each time step, as applying the same perturbation at all time steps would not cover the space of possible policies well (see Table~\ref{tab:policysearch}). In contrast, parameter-space exploration tends to go together with episode-based exploration, because certain parameters might only influence behavior in certain states, so such a perturbation has to be evaluated across multiple states to give a good indication of its merit~\citep{ruckstiess2010exploring}. However, exceptions to this pattern exist, especially in the area of exploration strategies of intermediate coherence. In the following paragraphs, papers presenting or analyzing specific exploration strategies will be discussed in more detail. We will start by discussing exploration strategies that apply explorative perturbations in the action space, before turning our attention to strategies that perturb the policy parameters. Finally, we will discuss the issue of what distribution these perturbations are sampled from.

\subsubsection{Action-space perturbing strategies }
\label{ssec:actionspace}
Using stochastic policies to generate action-space perturbations has been used at least since the early 80's.  \citet{barto1983neuronlike} proposed an early actor critic-type algorithm, that perturbed the output of a simple neural network before applying the thresholding activation function. This resulted in Bernouilli-distributed outputs \citep{williams1992simple}. 

Subsequent work also investigated the use of Gaussian noise in continuous action domains. \citet{gullapalli1990stochastic} introduced specific learning rules for the mean and standard deviation of Gaussian policies, which was introduced to learn policies with continuous actions.  \citet{williams1992simple} provided a more general algorithm that provides an update rule for a general class of policy functions including stochastic and deterministic operations. \citet{williams1992simple} notes that because a Gaussian distribution has separate parameters controlling location and scale, such random units have the potential to control both the degree of exploration as well as where to explore.  These early approaches all perturbed the action chosen at each time step independently.

While independent perturbation is still a popular method, attention has also turned to strategies that attempt to correlate behavior in subsequent time steps. An interesting early example can be found in \citet{morimoto2001acquisition}. That paper describes a hierarchical policy, where an upper-level policy sets a sub-goal which a lower-level policy then tries to achieve. Exploration on the upper-level by itself causes some consistency in the exploration behavior, as a perturbation in the goal-picking strategy will consistently perturb the system's behavior until the subgoal is reached. Additionally, the lower-level learner itself applies a low-pass filter on the action perturbations. As a result, similar perturbations will typically be applied on subsequent time steps. The authors applied this algorithm to learn a stand-up behavior for a real robot.\footnote{Other work has specifically evaluated the potential of lower-level sub-policies to decrease the diffusion time during the exploration process in the context of pure exploration \citep{machado2017laplacian}. This paper is explained in more detail in Sec. \ref{sec:pure-policy-seach}.
}

\cite{nachum2019does} studied hierachical methods in more detail, among others focusing on their exploration behavior. In their experiments, they investigate two simple exploration heuristics that share certain properties with hierarchical policies. The first heuristic, Explore \& Exploit, randomly sets 'goals' for a separately trained explore policies analogous to the high-level actions in a goal-conditioned hierarchical method. Goals stay active for one or multiple time steps. Their second method, Switching Ensembles, trains several separate networks that individually attempt to optimize rewards. During training, the active policy is periodically switched, and when the policies are different, this switching leads again to exploration behavior that is coherent over several time steps. \citeauthor{nachum2019does} find that both methods benefit from temporal coherence, and their results suggest that setting goals in a meaningful space might additionally benefit exploration. 

Similar to the approach by \citet{morimoto2001acquisition}, \citet{wawrzynski2015control} investigated performing explorative perturbations on physical robots. As the authors note, applying perturbations independently at each time step (e.g., independent draws from a stochastic policy) causes jerkiness in the trajectories, which damages the robot. As an alternative, the paper proposes to apply an auto correlated noise signal. This signal is generated in a slightly different way than the previously discussed approach, as it is generated by summing up independent perturbations from the last $M$ time steps. The authors explicitly evaluated the suggested strategy on various continuous robot control problems. Their experiments suggest that the proposed strategy leads to equivalent asymptotic performance (although sometimes a slower learning speed), while causing less stress to the robot's joints by reducing the jerkiness of trajectories.

Correlating the perturbations over several time steps, however, complicates the calculation of log-ratio policy gradients, as the policy is no longer Markov (as the selected action is no longer independent of earlier events given the state). \citet{lillicrap2016continuous} instead apply auto-correlated noise for their deep deterministic policy gradient (DDPG). Since DDPG is an off-policy algorithm, the generating distribution of the behavior policy does not need to be known, simplifying the use of various kinds of auto-correlated noise. They proposed generating this noise using an Ornstein-Uhlenbeck process, which generates noise with the same properties as that used by \citet{morimoto2001acquisition}. This paper did not focus on real-robot experiments, thus motor strain was not a major concern. They did, however, determine that auto-correlated noise does help learn in (simulated) `physical environments that have momentum', in other words, environments where a sequence of similar actions need to be performed to cause appreciable movement in high-inertia objects.

More recently, attention seems to have swung back to independent action perturbations, with recent work attempting to make the distribution from which actions are drawn more expressive, or the resulting explorative actions more informative. While classically, simple parametric distributions have been used as stochastic policies, these  typically cannot represent multi-modal policies. \citet{haarnoja2017reinforcement} point out that it is useful to maintain probability mass on \emph{all} good policies during the learning process, even if this results in a multi-modal distribution. In particular, a seemingly slightly-suboptimal mode might in a later stage of learning be discovered to actually be optimal, which would be hard to uncover if this mode was discarded earlier in the learning process. The authors define the exploration policy as maximizing an objective composed of a reward term and an entropy term. The solution to this maximization problem is an energy-based policy. As one cannot generally sample from such distributions, an approximating neural-network policy is fit to it instead. The authors show that with certain (initially) multi-modal reward distributions the method outperforms exploration using single-modal exploration policies. They also show empirically that a multi-modal policy learned on an initial task can provide a useful bias for exploring more refined tasks later.

Although maintaining a high entropy can be a useful strategy to obtain more informative data, it might be even more effective to directly maximize the amount of improvement to a target policy caused by data gathered using the exploratory behavior policy. This was the approach proposed by \citet{xu2018learning}, who study the optimization of the behavior policy in off-policy reinforcement learning methods, where the exploration policy can be fundamentally different from the target policy. The authors' insight is that good exploration policies might indeed be quite different from good target policies, and thus might not be centered on the current target policy, but instead have a separate parametrization. While the target policy is adapted in an off-policy manner in the direction of maximum reward, the behavior policy is separately updated by an on-policy algorithm towards greater improvements to the target policy. This is achieved by using an estimate of the improvement of the target policy as the reward of a `meta-MDP'. Experiments show that learned variance, and even more so a learned mean function, results in faster learning and better average rewards compared to conventional exploration strategies centered on the target policy. The authors attribute this performance gain to more diversity in the exploration samples leading to more global exploration.

\subsubsection{Parameter-space perturbing strategies }
\label{ssec:parameterspace}
Instead of using action-space perturbation, one might directly perturb the parameters of the policy. Especially where the parametrization of the policy can be restricted due to prior knowledge about the problem, it might be advantageous to do so. For example, if we know or assume that the optimal action will be directly proportional to the deviation from a set point, it is not informative to perturb the policy at the set point, because the policy will choose an action of 0 at the set-point regardless of the parameter value. This information is implicitly taken into account if parameters, rather than actions, are perturbed. 

Perturbing parameters rather than actions also ensure that any explorative action can also, in fact, be reproduced by some policy in hypothesis space \citep{deisenroth2013survey}. Referring back to the previous example, a non-zero action perturbation at the set-point, for example, would not be reproducible by any of the considered controllers.

Perturbing parameters also works very well together with temporally coherent exploration. A parameter vector might simply be perturbed only at the beginning of an episode, and then kept constant for the rest of the episode. Contrast this to action-perturbing schemes, where keeping the action perturbation constant for a whole episode in general would not yield behavior that covers the state-action space well.

The most straightforward way to find out what parameter perturbations improve a policy is to treat the entire interaction between the policy parameters and environment as a black-box system and calculate a finite-difference estimate of the policy gradient, keeping the policy perturbation fixed during the episode. An example of this approach is given by \citet{kohl2004policy}. For each policy roll-out, each policy parameter randomly chosen to be either left as is or to be perturbed by a adding a small positive or negative constant. After obtaining a set of such roll-outs, the policy gradient is then estimated using a finite-difference calculation. This method was demonstrated to able to optimize walking behaviors on four-legged robots better than earlier hand-tuned or learned gaits.

A potential problem with this approach is that stochasticity (from the policy or environmental transitions) can make gradient estimates extremely high-variance. In simulation, where stochasticity can be controlled, this can be addressed by using common random numbers, as was proposed by \citet{ng2000pegasus} in their PEGASUS algorithm to learn policies for gridworld problems and a bicycle simulator.  

Ratio-likelihood policy gradient estimators exploit the knowledge of the parametric form of the policy to calculate more informed estimates of the policy gradient. \citet{sehnke2010parameter} proposed a stochastic policy gradient with parameter-based exploration by positing a parameterized distribution $\pi(\theta|\rho)$ over the parameters $\theta$ of a (deterministic) low-level policy, and learning the hyper-parameters $\rho$. Their experiments showed that the resulting parameter-perturbing, episode-based exploration strategy outperformed conventional action-perturbing strategies on several simulated dynamical systems tasks, including robotic locomotion and manipulation tasks. \citet{ruckstiess2010exploring} extended the idea of parameter-based exploration to several other policy-search algorithms, including a new method called state-dependent exploration. In that approach, perturbations are defined in the action-space, but are generated based on an `exploration function' that is a deterministic function of a randomly generated vector and the current state. The authors show that state-dependent exploration is equal to parameter-based exploration in the special case of linear policies, and argue that in other cases it combines some of the advantages of parameter-based and action-based exploration. The resulting data was then used to perform REINFORCE updates. As these updates do not fully account for the dependency between actions, they might thus have an increased variance.

\citet{theodorou2010generalized} proposed `per-basis' exploration, which is a variant parameter-based exploration scheme where the perturbation is only applied to the parameter corresponding to the basis function with the highest activation and kept constant as long as that basis function had the highest activation. \citet{theodorou2010generalized} noted that they emperically observed this trick improved the learning speed.

The effect of step-based (independent) versus correlated or episode-based exploration was further studied by \citet{stulp2012path}. 
They investigated connections between CMA-ES (Covariance matrix adaptation evolutionary strategies) from the stochastic search literature \citep{hansen2001completely} and PI$^2$, a reinforcement learning algorithm with roots in the control community \citep{theodorou2010generalized}. \citet{stulp2012path} found that episode-based exploration indeed outperformed per time-step exploration on a simulated reaching task. Furthermore, it also outperformed per-basis exploration proposed by \citet{theodorou2010generalized}. 

\cite{salimans2017evolution} applied the idea of episode-level log-ratio policy gradients (as used in earlier work by e.g. \citealt{sehnke2010parameter}) to complex neural network policies. They proposed the use of virtual batch normalization to increase the methods sensitivity to small initial difference. The authors connect this method to approaches from the evolutionary strategies community (e.g. \citealt{koutnik2010evolving}). \cite{salimans2017evolution} found their approach to compare favorably to Trust Region Policy Optimization (TRPO, \citealt{schulman2015trust}), an action-space perturbing method, on simulated robotic tasks. Furthermore, the approach performed competitively with the Asynchronous Advantage Actor Critic (A3C, \citealt{mnih2016asynchronous}), while training an order of magnitude faster.

\citet{conti2018improving} combined similar ideas from the evolutionary strategies community with directed exploration strategies. This combination results in two new hybrid approaches, where evolutionary strategies are used to maximize a scalarization of the original reward-maximization objective with a term encoding novelty and diversity. A heuristic strategy for adapting the scalarization constant is proposed. The proposed approach is evaluated on a simulated locomotion task and a benchmark of Atari games, where it outperforms a regular evolutionary strategy approach and performs competitively with the deep RL approach by \citeauthor{fortunato2018noisy} (\citeyear{fortunato2018noisy}; described below). In this experiment, the evolutionary strategies were given more frames, but still ran faster due to better parallelizability.

\Citet{hoof2017generalized} apply auto-correlated noise in parameter space. This noise is  distributed similarly to that proposed by \citet{morimoto2001acquisition} and \citet{lillicrap2016continuous}, but applied to the parameters rather than the actions. As a result, intermediate trade-offs between independent and episode-based perturbations are obtained. The latent parameter vector violates the independence assumption of step-based log-ratio policy gradients methods, which is resolved by explicitly using the joint log-probability of the entire sequence of actions. Expressing this log-probability in closed form requires the use of a restricted policy class (e.g., linear policies). Note that the auto-correlated action-space noise used by \cite{lillicrap2016continuous} was applied in an off-policy setting, avoiding this problem. Auto-correlated parameter-space noise was compared to several baselines, including action-space perturbations as well as episode-based and independent parameter-space perturbations. On various simulated and real continuous control tasks, intermediate trade-offs between independent and episode-based perturbation led to faster learning and a way to control state space coverage and motor strain.

Two methods \citep{plappert2018parameter, fortunato2018noisy}  independently proposed strategies to use parameter-based perturbations for reinforcement learning approaches based on deep neural networks. Both of these works consider both value-networks and policy-based approaches. Here, we will discuss the policy-based variants.  Both of the approaches  also build on the principle of episode-based perturbation of the parameter space, but better exploit the temporal structure of roll-outs than previous studies \citep{sehnke2010parameter, salimans2017evolution} that largely ignored it. 

By making the perturbations fixed over an entire episode and using the reparametrization trick, these methods allow the use of a wide range of policies, but possibly increase variance. Subsequent actions are now conditioned on a shared sample from the noise distribution, and their conditional independence means the trajectory likelihood can be factored as usual. As a result, this methods are applicable to non-linear neural network policies. \citet{plappert2018parameter} used a pre-determined amount of noise that was decreased over time according to a pre-set schedule. On the other hand, \citet{fortunato2018noisy} learned the magnitude of the parameter noise together with the policy mean

Applying this principle in an off-policy setting is relatively easy: since any behavior policy could be used to generate data, this could easily be the current deterministic target policy with noise added to the parameters. \cite{plappert2018parameter} modified the Deep Deterministic Policy Gradient (DDPG, \citealt{lillicrap2016continuous}) algorithm in this manner. With on-policy algorithms, this is more challenging. On-policy algorithms with per time-step updates tend to also require stochastic action selection in each time step. For the on-policy Asynchronous Advantage Actor Critic (A3C, \citealt{mnih2016asynchronous}) in \citet{fortunato2018noisy}; and for Trust Region Policy Optimization (TRPO, \citealt{schulman2015trust}) in \citet{plappert2018parameter}; this problem was resolved by using a combination of parameter perturbations and stochastic action selection.

The NoisyNet-A3C method by \citet{fortunato2018noisy} compared favorably to the baseline A3C variant on a majority of 57 Atari games.
The parameter-exploring variants of DDPG and TRPO proposed by 
\citet{plappert2018parameter} compared favorably to baselines with (correlated or uncorrelated) action-space noise on several simulated robot environment. 
\footnote{\citet{colas2018gep}, whose exploration method is discussed in Sec. \ref{sec:pure}, verify the performance difference between action- and parameter space noise for DDPG, and compare their methods to the exploration methods by \citet{lillicrap2016continuous} and \citet{plappert2018parameter}. Like these methods, their proposal benefits from the flexibility of DDPG as off-policy method to work with data from any behavior policy.}
 
Incoherent behavior causes particular problems in multi-agent learning, as pointed out by \citet{mahajan2019maven}. In cooperative decentralized execution scenarios, uncoordinated exploration between the agent can lead to state visitation frequencies for which the factorized q-function approximation is catastrophically bad. Such failure traps the agents in suboptimal behavior. \citet{mahajan2019maven} remedy the situation by making exploration at train time coherent across both time and the individual agents, by conditioning on a common latent variable generated by a high-level policy\footnote{Note that although the perturbed parameters are of a value network, this hybrid value- and policy based approach was covered here as the novelty in the exploration strategy stems from the high-level parametrized policy.}. A separate variational network is used to estimate a mutual information term which avoids collapse of the high-level policy on constant low-level behavior. The proposed approach is compared on both a toy task as well as challenging scenarios from the StarCraft Multi-Agent Challenge \citep{samvelyan2019starcraft}. 

\subsubsection{The distribution of perturbations}
\label{ssec:policydist}
Separate from the issue of how perturbations are applied, is the issue of what distribution these perturbations are sampled from. Often, these are Gaussian distributions centered on the policy mean, leaving the choice of standard deviation open. Other parametric policies may have different parameters controlling the amount of exploration.
Sometimes, such parameters are treated as additional hyperparameters \citep{silver2014deterministic} or governed by a specific heuristic \citep{gullapalli1990stochastic}.
More commonly, they can also be adapted like the other parameters during learning. The most straightforward way is to adapt the parameters controlling this standard deviation using the same policy gradient \citep{williams1992simple}. Without additional regularization, policy gradient methods will tend to reduce the uncertainty, leading to a loss of exploration that is hard to control and might result in premature convergence to a suboptimal solution \citep{williams1991function, peters2010relative}.   

To address this problem, several approaches of them have been proposed. Many of them involve regularization using the entropy of the policy or the relative entropy from a reference policy. These entropies can either be constrained or added as regularization term to the optimization objective. A unified view on regularized MDPs is presented by \cite{neu2017unified, geist2019theory}.

As an example, \cite{bagnell2003covariant} studied   natural policy gradients through the lens of limiting the divergence between successive policies. They found the natural gradient can be derived from a bound on the approximate Kullback-Leibler divergence between trajectory distributions. 
Dynamic policy programming \citep{azar2011dynamic} uses a similar formulation but instead uses the relative entropy as penalty term. 
\cite{schulman2015trust} provides a more exact method, focused on limiting the equivalent expected Kullback-Leibler divergence between policies. They  connects this update to the idea of trust region optimization, and provides several steps to make scale this type of network to deep reinforcement learning architectures with tens of thousands of parameters. 

The `relative entropy policy search' method proposed by \cite{peters2010relative} bounds the KL divergence in the joint state-action distribution to avoid a loss of exploration during training. Their bound on the joint KL is a stricter condition than a bound on the expected KL divergence of state-action conditionals which has theoretically attractive properties \citep{zimin2013online}. However, the method is more complex and seems harder to scale to deep architectures \citep{duan2016benchmarking}. \citet{bas2020logistic} propose a method building on relative entropy policy search, that combines regularization terms on the joint- and expected KL. A large benefit is that the resulting algorithm can be faithfully implemented in deep RL frameworks, and thus does need further approximations of the policy. 

An alternative used early on was to add the derivative of the policy entropy to the policy updates. \cite{williams1991function} found this strategy to improve exploration open on a toy example and several optimization problems. This strategy has also proved fruitful in practice in deep learning approaches:  \cite{mnih2016asynchronous} applied this strategy and informally observed it lead to improved exploration by discouraging premature convergence. Such an entropy term can be seen as a special case of the expected relative entropy objective, with the reference policy being the maximum entropy distribution. \citet{neu2017unified} studied such regularized objectives in detail, and conclude that a entropy penalty based on samples from the previous policy distribution distribution can lead to optimization problems.

The entropy regularization methods in the previous paragraph only took the instantaneous policy entropy into account. \cite{haarnoja2017reinforcement,haarnoja2018soft} instead propose two methods that explicitly encourages policies that reaching high-entropy states in the future. They note their method improves exploration by acquiring diverse behaviors. 

\section{Bonus-Based/Optimism-Based Exploration}\label{Bonus}
A popular category of exploration methods commonly used in domains with weak or sparse explicit reward structure is the bonus-based methods. In this category, extrinsic reward $r(s, a)$ is augmented with a \textit{bonus} term that is often demonstrated as a form of intrinsic reward \citep{oudeyer2007intrinsic} to encourage better exploration. The term bonus was first introduced in the early nineties by \cite{sutton1990integrated} in the tabular setting. Algorithms that adopt the bonus-based exploration approach employ different bonus calculation techniques to encourage the choice of action that leads to a higher level of uncertainty and consequently, novel or informative states. 

In an environment with underlying MDP $\mathcal{M}$, upon selection of the state-action pair $(s, a)$, the explicit reward  $r(s, a)$ is observed by the RL agent, and the bonus term $\mathcal{B}(s, a)$ is computed by the RL agent. Thus, the total reward obtained by the agent by taking action $a$ at state $s$ is defined as, 
\begin{align}
    r^+(\textbf{s},\textbf{a}) := r(\textbf{s},\textbf{a}) \oplus \mathcal{B}(\textbf{s},\textbf{a}),
\end{align}
where the operator $\oplus$ denotes the aggregation between the two sources of environment (extrinsic) reward and bonus term.

In designing bonus-based exploration algorithms, two main questions arise: 
\begin{enumerate}
    \item How should the bonus function $\mathcal{B}(s,a)$ be specified to yield an effective exploration behavior?
    \item How should we combine the two separately acquired sources of information, exploration bonus denoted by $\mathcal{B}(s,a)$ and \textit{extrinsic} reward denoted by $r(s,a)$?
\end{enumerate}

Exploration methods described in this section adopt three different approaches to compute the additive bonus term, namely \textit{optimism-based}, \textit{count-based}, and \textit{prediction error-based}. In the optimism-based methods, the bonus term is implicitly embedded in the value function initial value estimates. In the count-based methods, where the novel state-action pairs are the ones that are less frequently visited, the bonus term is calculated based on some form of state-action visitation counts. Another way of calculating the bonus term is through a prediction model of environment dynamics and measuring the induced prediction error. The interplay between the optimism-based approach and the count-based method in designing the bonus term is subtle, and it is often hard to distinguish between these two paradigms. Thus, in the optimism section, we only discuss the optimism-based approaches that do not explicitly use any notion of count in their choice of bonus. In the count-based section, we only discuss the methods that explicitly employ a notion of count in their bonus definitions. In this survey, we choose to interpret the optimistic initialization heuristic as a type of bonus assigned to unseen state-action pairs to encourage exploration.

All these three categories are divided into two sub-categories of \textit{tabular} and \textit{function approximation} methods. In the tabular setting, the state and action spaces are small enough that the value function estimate can be presented as a lookup table. In the function approximation setting, on the other hand, due to the infinite or large nature of state and action pairs, the value function is represented as a parameterized function rather than a table. At the beginning of each section, we provide a table that summarises the papers discussed.

\subsection{Optimism-based methods}
Optimistic exploration is a category of exploration methods that adopt the philosophy of Optimism in the Face of Uncertainty (OFU) principle, which was first introduced as an ad-hoc technique by \cite{kaelbling1996reinforcement,kaelbling1993learning,sutton1991dyna,thrun1992active}. From an optimistic perspective, a state is considered a good state if it induces a higher level of uncertainty in the state-value estimate and greater return potential. Optimistic exploration methods are typically realized by implicitly utilizing an \textit{exploration bonus} either in the form of \textit{optimistic initialization} \citep{even2002convergence, sutton1998reinforcement, szita2008many} or \textit{Upper Confidence Bounds} (UCBs)\citep{strehl2008analysis, jaksch2010near}. In the optimistic initialization approach, the key assumption is that the unvisited state-action pairs yield the best outcome, whereas in the UCB-based methods, the unvisited state-action pairs are assumed to collect the outcome proportional to the largest statistically possible reward. In this section, we only focus on the methods that do not employ count-based approaches to implement the OFU principle as we address the count-based methods in a separate section.

\begin{table} 
\centering
\begin{tabular}{p{6cm}|p{4.5cm}|p{4.5cm}}
    Reference & Approach & Performance measure \\
    \hline
    &\textbf{Tabular Methods}& \\
    \hline
    \cite{brafman2002r} & optimistic initialization & PAC bound \\
    \cite{auer2007logarithmic} & UCB-based & PAC bound\\
    \cite{szita2008many} & optimistic initialization & PAC bound/Empirical-Grid and Empirical-Toy MDP\\
    \hline
    &\textbf{Function Approximation Methods}& \\
    \hline
    \cite{jong2007model} & optimistic initialization & Empirical-Grid world/Mountain Car\\
    \cite{nouri2009multi} & optimistic initialization & PAC bound/Empirical \\
    \cite{ortner2012online} & UCB action selection & Regret bound \\
    \cite{kumaraswamy2018context} & UCB action selection & Empirical-MuJoCo and Puddle World/ PAC Bound\\
    \cite{ciosek2019better} & optimistic initialization & Empirical-MuJoCo\\
    \cite{Rashid2020Optimistic} & UCB action selection &  Regret Bound/Empirical-Atari\\
    \cite{seyde2020learning} & UCB action selection & Empirical-MuJoCo
\end{tabular}
\caption{Exploration methods that implement an optimism-based bonus mechanism.}
\label{tab:optimism-based}
\end{table}

\subsubsection{Optimism-based methods: tabular}\label{sec:optim-tab}

A model-based approach proposed as a generalization of $E^3$ algorithm \citep{kearns2002near} (discussed in detail in Section \ref{sec:count-based-tabular}) is the \textit{R-max} algorithm \citep{brafman2002r}, which models agents' interactions in the context of zero-sum \textit{stochastic games} (SG) instead of MDPs. In R-max \citep{brafman2002r}, the agent always maintains maximum likelihood estimates of environment dynamics and reward function if the observed data is sufficiently rich. The algorithm uses the approximate model estimates for a state-action pair if its visitation count exceeds a certain threshold. The optimistic approach is adopted at the initialization phase of the model, where all actions in all states are assumed to return maximum reward $\Rmax$. R-max benefits from a built-in mechanism for resolving the exploration vs. exploitation dilemma because of the model estimation and optimism. That is, taking the optimal action according to the learned model results in either exploring a previously unknown state or obtaining the near-optimal reward. In R-max, a state is marked as \textit{known} if the number of states reachable (based on the learned model) from that state passes a fixed threshold; therefore, it is no longer considered a novel state. The sample complexity analysis conducted by \cite{brafman2002r} shows near-optimal performance in a polynomial number of time-steps (assuming the state-space is finite). The R-max algorithm also attains near-optimal polynomial time average reward in $|\states|, |\actions|$ and mixing time $T$. 

In a similar approach, authors in \cite{szita2008many} propose a new sample efficient and model-based exploration algorithm called Optimistic Initial Model (OIM) in an MDP framework with finite state and action spaces. The proposed method assigns an optimistic value to unknown areas, and if the sampled state is among the set of 'Garden of Eden' states, it receives the maximum reward of $\Rmax$. The RL agent in this method builds an approximate model of environment dynamics and updates the value functions using dynamic programming. To handle the full update sweep complexity imposed by dynamic programming, authors adopt the prioritized sweeping algorithm of \cite{wiering1998efficient}. Theoretically, the authors show almost sure convergence of the proposed method to a near-optimal policy in polynomial time under a lower-bound assumption on $\Rmax$. Experimentally, OMI's performance is assessed against $\epsilon$-greedy, Boltzmann and some other exploration methods in three environments of \textit{RiverSwim} and \textit{SixArms} \citep{strehl2006pac}, \textit{Maze} \citep{wiering1998efficient} and \textit{Chain}, \textit{Loop} and \textit{FlagMaze} \citep{meuleau1999exploration,strens2000bayesian,dearden1999model}.

The seminal idea of employing optimistic \textit{upper confidence bound} (UCB) to encourage optimistic exploration policies in RL setting was first introduced by \citet{strehl2004exploration}. \cite{auer2007logarithmic} proposed UCRL algorithm as the first near-optimal exploration method that extends the idea of optimistic UCBs in RL. UCRL computes a count-based upper bound on the empirical estimates of reward and transition probability after each visit and then switches between policies based on the observed gap and calculates and a new policy. Later, \cite{jaksch2010near} proposed UCRL2 as an extension to \cite{auer2007logarithmic}. Apart from optimistic initialization, both UCRL and URCL2 implement count-based UCBs to encourage exploration. Thus, we postpone further explanation regarding these methods to the count-based section.

\subsubsection{Optimism-based methods: function approximation}

After the successful application of OFU to RL with finite state-action MDPs, which we addressed in Section \ref{sec:optim-tab}, some recent approaches extended this idea to MDPs with large or infinite state-action spaces \citep{azar2017minimax, bartlett2012regal, fruit2018efficient, filippi2010optimism, jaksch2010near, tossou2019near}. This section provides a comprehensive overview of the methods that employ OFU in the  MDPs with large or infinite state-action spaces.

The study presented by \cite{jong2007model} is a model-based exploration-exploitation trade-off algorithm for continuous state spaces, termed Fitted R-max. The proposed algorithm is a combination of R-max \citep{brafman2002r} with fitted value iteration \citep{gordon1995stable}. The algorithm first updates environment models at each time-step and then applies the value iteration step to solve their proposed Bellman optimality equation. In discrete setting, the proposed method simply implements the optimistic value function proposed by  \cite{brafman2002r} and control the exploration-exploitation trade-off through a visitation count threshold. In continuous setting, the authors propose a new counting method based on the sum of the unnormalized kernel values based in the estimated environment dynamics. The performance of the Fitted R-max algorithm is experimentally tested in the two environments Mountain Car \citep{sutton1998reinforcement} and Puddle World \citep{kearns2002near}. 

Another line of work that implements the OFU principle in continuous state space environments is \citet{nouri2009multi}, where the authors combine their proposed Multi-resolution Exploration (MRE) algorithm with fitted Q-iteration \citep{antos2008fitted}. Their proposed model-based method is built upon \citet{kakade2003exploration} and introduces a method that measures the uncertainty associated with the visited states through building regression trees, termed as \textit{knownness-tree}. Knownness-tree is used to model the environment transition dynamics and optimistically update its model at each time step. Theoretically, \cite{nouri2009multi} show, under some smoothness assumption on transition dynamics, the near-optimal performance of the proposed algorithm, and assess the performance of MRE against $\epsilon$-greedy algorithm empirically in continuous Mountain Car environment \citep{sutton1998introduction}. 

\cite{kumaraswamy2018context} propose a model-free computationally efficient exploration strategy based upon computing Upper-Confidence Least-Squares (UCLS), which are UCBs for least-squares temporal difference learning (LSTD). Since LSTD maintains the agent's past interactions efficiently, the computed upper confidence bounds induce context-dependent variance, which encourages the exploration of states with higher variance. This study provides the first theoretical results that obtain UCBs for policy evaluation using function approximation. Empirically, UCLS algorithm shows outperformance over DGPQ \citep{grande2014sample}, UCBootstrap \citep{white2010interval}, and RLSVI \citep{osband2016generalization} in Sparse Mountain Car, Puddle World and RiverSwim environments.

\cite{ciosek2019better} provide an optimistic actor-critic algorithm to resolve two phenomena: \textit{pessimistic under-exploration}, that is deviating the algorithm from sampling actions that result in improvement on the critic estimates, and \textit{directionally uninformed action sampling}, which is uniform sampling of actions that are lying in two opposite sides of mean in Gaussian policies. They state that these two phenomena prevent state-of-the-art actor-critic-based algorithms such as SAC \citep{haarnoja2018soft} from performing efficient exploration. To tackle these challenges, they calculate approximate upper confidence bounds on the value function estimates to encourage directed exploration and lower confidence bound to prevent overestimation of the value function estimates. They benchmark their proposed algorithm (OAC) in high-dimensional MuJoCo tasks, and the plotted results demonstrate marginal improvement against the SAC algorithm.

To avoid the pessimistic initialization phenomenon, commonly used in deep network initialization schemes, \cite{Rashid2020Optimistic} propose an optimistic initialization algorithm, termed Optimistic Pessimistically Initialised Q-Learning (OPIQ) that decouples optimistic initialization of Q function from network initialization. \cite{Rashid2020Optimistic} propose a simple count-based bonus augmented with the Q-value estimates. In the tabular setting, their proposed algorithm is based on \cite{jin2018q}. In the Deep RL setting, they adopt commonly employed methods of calculating pseudo counts such as \cite{bellemare2016unifying,ostrovski2017count} to compute the additive bonus term. OPIQ is evaluated in the three domains toy randomized Markov chain, Maze and Montezuma's Revenge against the naive extension of UCB-H \citep{jin2018q} to the deep RL and some variations of DQN augmented with some state-of-the-art pseudo-count estimate methods.

When both the environment dynamics and task objective are unknown to the RL agent, \cite{seyde2020learning} propose a model-based exploration algorithm, termed Deep Optimistic Value Exploration (DOVE), to encourage deep exploration through adopting optimistic value function. Throughout each episode, DOVE learns a transition function and reward function using supervised learning. The initial conditions are applied to the learning policy generated by perturbing locally observed states fetched from the replay memory. The local perturbation performed throughout each episode is employed to ensure information propagation. Empirically, \citet{seyde2020learning} benchmark DOVE in some high-dimensional continuous control MujuCo tasks.

\subsection{Count-based bonus}
One way to model the intrinsic reward is to measure how surprising a state-action pair is. An intuitive approach to measuring surprise is to count how frequently a particular state-action pair is visited.  In the count-based setting, the notion of bonus is defined as a function of state-action pair visitation count. In table \ref{tab:visitation-count}, we provide a list of exploration methods that explicitly employ a form of visitation count to control the exploration-exploitation trade-off.  
\begin{table} 
\centering
\begin{tabular}{p{6cm}|p{5cm}|p{5cm}}
    Reference & Bonus Type & Performance measure \\
    \hline
    &\textbf{Tabular Methods}& \\
    \hline
    \cite{sutton1991integrated} &  count-based & Empirical-Grid world \\
    \cite{moore1993prioritized}  &  count-based threshold/optimistic initialization & Empirical-Grid world \\
    \cite{kaelbling1993learning} &  count-based &  Empirical-Toy MDP\\
    \cite{dayan1996exploration} & count-based  & Empirical-Grid world\\
    \cite{tadepalli1998model} & count-based/optimistic initialization  &  Empirical-Grid world/AGV-scheduling\\
    \cite{kearns2002near} &  count-based/optimistic initialization  & PAC bound \\
    \cite{kakade2003exploration} & distance-based count & PAC bound\\
    \cite{strehl2006pac} & UCB-based & PAC bound\\
    \cite{auer2007logarithmic} & UCB-based & PAC bound\\
    \cite{jaksch2010near} & UCB-based & PAC/Regret bound\\
    \cite{azar2017minimax} & UCB-based & Regret bound\\
    \cite{kolter2009near} & count-based  & PAC bound\\
    \cite{pazis2013pac} & optimistic initialization / distance-based bonus  & PAC bound/Empirical \\
    \cite{guo2015concurrent} & count-based threshold &  PAC bound\\
    \cite{jin2018q} & count-based/UCB-based  & PAC bound \\
    \cite{Wang2020Q-learning} & count-based/UCB-based  & PAC bound\\
    \hline
    &\textbf{Function Approximation Methods}& \\
    \hline
    \cite{} & & \\
    \cite{bellemare2016unifying} & density-based  & Empirical \\
    \cite{fu2017ex2} & density-based  & Empirical\\
    \cite{martin2017count} & density-based  & Empirical\\
    \cite{tang2017exploration} & count-based  & Empirical\\
    \cite{machado2018count} &count-based  & Empirical\\
    \cite{Rashid2020Optimistic} & optimistic initialization/ count-based  &  Empirical
\end{tabular}
\caption{Count-based methods.}
\label{tab:visitation-count}
\end{table}

\subsubsection{Count-based bonus: tabular} \label{sec:count-based-tabular}

In the tabular setting, counting visited states or state-action pairs is a trivial problem and the bonus term is typically used in one of the following forms, 
\begin{align}\label{eq:bonus-forms-tabular}
   \mathcal{B}(\textbf{s},\textbf{a})~\text{or}~\mathcal{B}(\textbf{s}) \propto  
\begin{cases}
\sqrt{\frac{\ln{n}}{N(\textbf{s},\textbf{a})}}~\text{or}~\sqrt{\frac{\ln{n}}{N(\textbf{s})}},\\    
\frac{1}{\sqrt{N(\textbf{s},\textbf{a})}}~\text{or}~\frac{1}{\sqrt{N(\textbf{s})}},\\
\frac{1}{N(\textbf{s},\textbf{a})}~\text{or}~\frac{1}{N(\textbf{s})}.
\end{cases}
\end{align}
where $n$ denotes the total number of time-steps taken by the RL agent.

An intuitive interpretation of equation \eqref{eq:bonus-forms-tabular} is that the bonus for visiting a state-action pair $(s,a)$ is highest when $(s,a)$ is novel, and decays each time the pair $(s,a)$ is revisited. The main limitation of such definitions of bonus is that they are mainly applicable in tabular settings, where the set of state-action pairs is countable and finite. Although bonus-based methods employed in the tabular settings are not necessarily suitable for large state-and-action space settings, they still provide useful intuitions. The first study that employed the notion of bonus in the context of exploration algorithms was \cite{sutton1991integrated}, where a new RL architecture Dyna-Q+ was proposed. Dyna-Q+, which is a combination of Dyna architecture \citep{sutton1991dyna} and Watkins Q-learning \citep{watkins1992q} uses an additional explicit count-based exploration bonus assigned to state-action pair $(s,a)$. The bonus term used in Dyna-Q+ is proportional to the square root of the number of time steps that have elapsed after the last trial of action $a$ at state $s$. This exploration bonus is added to the update rule designed to update Q-value. The main advantage of using such bonuses is to increase the chance of visiting the state-action pairs that have not been frequently visited. \cite{sutton1991dyna} tests his proposed model in the two RL environments Blocking and Shortcut tasks, designed as a small $2D$ maze environment against two other variations of Dyna-Q that do not use exploration bonus to encourage exploration. \cite{sutton1991dyna} shows that in both experiments, Dyna-Q+ outperforms other variations of Dyna-Q in the setting, where the performance is measured with respect to the collected reward. To address the two issues, namely high computational expense raised by \cite{kaelbling1993learning}, and instability of the bonus raised by \cite{sutton1991dyna} in large state and action spaces, \cite{moore1993prioritized} proposed the \textit{prioritized sweeping} algorithm that uses a preset threshold parameter $T_{board}$ to determine whether the state-action pair is worth exploring more or not. Prior to reaching the visitation threshold, the bonus parameter is set to the max return value in the discounted reward setting $\Rmax/(1-\gamma)$. Once the visitation count exceeds the optimistic threshold, the algorithm uses the non-optimistic true discounted return. Apart from the tabular nature of this approach, its main bottleneck is that the key hyperparameter $T_{board}$ is prefixed and set manually. The proposed algorithm's performance is experimentally tested against Dyna-Q in two deterministic and stochastic grid world environments.

An early study that adopts the OFU principle in a model-based setting for exploration is \cite{tadepalli1998model}. The proposed \textit{H-learning} algorithm is designed in the context of the \textit{average-reward RL} setting, where the RL agent's goal at each time-step $t$ is to optimize the average expected reward. At each time-step $t$, an empirical model of the environment is computed, and consequently, a set of greedy actions accessible from the current state with respect to the Bellman equation for average-reward RL is reported, as well as the expected long-term advantage function $h(s)$. The long-term advantage function $h(s)$ reflects the long-term impact of start state $s$ on the obtained expected average reward. Eventually, the final expected average reward associated with actions in the set of greedy actions at the current state $s$ is calculated with respect to $h(s)$, a temperature parameter $\alpha$ and expected average reward computed at time $t-1$. Experimentally, the proposed H-learning algorithm is compared with four other exploration methods, random exploration, counter-based exploration, Boltzmann exploration, and recency-based exploration in a two-dimensional grid world with discrete state and action spaces, termed Delivery domain. The non-tabular version of the H-learning algorithm is proposed based on local linear regression as the function approximator and Bayesian network representation of the action space. The extended H-learning method called \textit{LBH-learning} is tested in three AGV-scheduling tasks \citep{maxwell1982design} and compared to six different H-learning baselines.

The \textit{Explicit Explore or Exploit} algorithm (known as $E^3$) \citep{kearns2002near} adopts a model-based approach that initiates the exploration phase by dividing the set of states into two categories of known and unknown states. A state is considered to be known if the number of state visitations passes a certain threshold such that the learned dynamics are sufficiently close to the true one. If the current state is unknown, the algorithm calls the procedure of \textit{balanced wandering}, in which the algorithm chooses the least frequent action at the unknown state and assigns the max reward to the unknown state. When the algorithm is not engaged in the balanced wandering phase, it performs two offline optimal policy computations sub-routines. Later, \cite{kakade2003exploration} proposed Metric-$E^3$ algorithm as a generalization of $E^3 algorithm$. Metric-$E^3$ provides the time complexity bound on finding a near-optimal policy that depends on the covering numbers of the state space rather than the size of the state space as presented by \cite{kearns2002near}. This difference is mainly due to the difference in their definition of a "known" state.  

In the context of undiscounted RL, \cite{auer2007logarithmic} use the notion of upper confidence bounds to manage exploration-exploitation trade-off. In their study, count-based confidence bound proportional to $\sqrt{\frac{1}{N(s,a)}}$ is updated at each step and, together with empirical estimates of reward and transition functions, help the agent to control the exploration-exploitation trade-off. The regret analysis performed by \cite{auer2007logarithmic} shows logarithmic performance in the number of time steps taken by the algorithm based on the optimal policy. 

\cite{kolter2009near} provide an explicit notion of bonus, called \textit{Bayesian Exploration Bonus}, to manage exploration-exploitation trade-off. Their proposed algorithm focuses on the Bayes-adaptive RL setting with a tabular representation of state-action space. The bonus term is proportional to $\frac{1}{1+N(s,a)}$, where $N(s,a)$ is calculated based on the number of visitation counts implied by the prior. \cite{kolter2009near} provide a template for count-based bonus terms in the form of a theorem stating that any algorithm that adopts an exploration bonus of the form $\frac{1}{(N(s,a))^p}$ with $p \leq 1/2$ is not a PAC-MDP. In their proposed algorithm, called BEB, the action-selection is performed with respect to the mean of the current learned belief over transition model, with an additional Bayesian bonus. In the main theorem of this work, the authors show that their proposed algorithm, while allowing a higher rate of exploration, provides a near-optimal sample complexity bound, which is polynomial in $|\states|$, $|\actions|$, and time horizon $T$, where the optimality is defined in the Bayesian sense.

In the continuous state space setting, \cite{pazis2013pac} introduce C-PACE as a PAC-optimal exploration algorithm for continuous state MDPs. C-PACE adopts the OFU principle in the estimation of the Bellman equation. At each time step, from the k-nearest neighbours, the action that maximizes the optimistic Q value function is selected. The optimistic Q value function is defined based on the knowledge of k-neighbouring state-action pairs (the bonus term) and the immediate reward obtained upon transiting to any neighbouring pairs. C-PACE assumes the existence of a Lipschitz continuous distance metric over the set of state-action pairs. The main result of this paper provides a PAC bound that shows the near-optimal C-PACE performance with respect to the covering number of the state-action space. Finally, the authors evaluate the performance of C-PACE in a simulated HIV treatment environment.

\cite{guo2015concurrent} propose a confidence-based exploration algorithm called PAC-EXPLORE in a model-based setting, which is operationally very similar to the $E^3$ algorithm \citep{kearns2002near}, with the difference in the confidence bounds used to compute policies that are practically more efficient. The PAC-EXPLORE algorithm takes a state-action pair visitation threshold and divides the space of state-action pairs into two clusters of known and unknown pairs. If the state falls into the set of known states, the algorithm applies the same technique as in \cite{wiering1999explorations} to estimate confidence bounds on transition probability. Authors in this work show that by sharing the experience of concurrently running agents on top of \cite{wiering1999explorations}, one can achieve linear improvement on the algorithm's sample complexity.

Delayed Q-learning \citep{strehl2006pac} is one of the first papers that study model-free PAC optimal algorithm. At each time-step $t$, the agent keeps track of three values for each visited state-action pairs $(s,a)$, the value function $Q_t(s,a)$, the Boolean flag $\operatorname{LEARN}_t(s,a)$ that indicates whether or not the change has occurred to the Q estimate for $(s,a)$, and a visitation counter $N(s,a)$. The exploration-exploitation trade-off is managed based on a visitation count threshold and the value $\operatorname{LEARN}_t(s,a)$. When the visitation count for $(s,a)$ is larger than a pre-set threshold and the $\operatorname{LEARN}_t(s,a)$ is true, the $Q_t(s,a)$ estimate is updated. At the initial phase, the Boolean flag $\operatorname{LEARN}(s,a)$ is set to TRUE for all state-action pairs and $N(s,a)$ is set to zero. \cite{strehl2006pac} under certain assumptions prove that their proposed algorithm is PAC-MDP in the tabular setting.

\cite{jin2018q} provide two types of upper confidence-based bonuses for Q-learning in the episodic tabular MDP setting: 1) Hoeffding-style bonus, 2) Bernstein-style bonus. By employing the Hoeffding-style bonus, the authors show $\mathcal{O}(\sqrt{T})$ regret dependency with respect to the total number of time-step $T$. They also show $\sqrt{H}$ improvement by using Bernstein-style bonus over the Hoeffding-style bonus exploration algorithm, where $T$ denotes the time horizon.

\cite{Wang2020Q-learning} introduce another method that addresses the sample efficiency of model-free algorithms by adopting UCB-exploration bonus in Q-learning.  Their proposed UCB-based algorithm maintains two types of visitation counts, $N_t(s,a)$ that denotes the number of times the pair $(s,a)$ has been visited up to time-step $t$, and $\tau(s,a,k)$ that records the number of time steps that state-action pair $(s,a)$ has occurred for the $k$-th time. If $(s,a)$ has not been visited $k$ times, then $\tau(s,a,k) = \infty$. At each time-step, a bonus term proportional to $\sqrt{\frac{|\states||\actions|\ln(N_t(s,a))}{N_t(s,a)}}$ is added to the discounted value estimate, and the action-value function gets updated accordingly. To assess the PAC efficiency of the proposed algorithm, the authors first propose a learning instance illustrating $\Omega(1/\epsilon^3)$ lower bound incurred by Delayed Q-learning, which leaves a quadratic gap in $1/\epsilon$ from the best known lower bound in the class of UCB-based exploration algorithms. 

\subsubsection{Count-based: function approximation}
Despite the near-optimal performance guarantees often achieved in the tabular setting, these methods are mostly not suitable for environments with large or infinite state spaces. This section summarizes exploration methods that adopt a notion of visitation count to design exploration algorithms for environments with large state and action spaces. 

\cite{bellemare2016unifying} revisit the problem of extending count-based exploration to non-tabular setting and propose a density model that hinges upon ideas from the intrinsic motivation literature (refer to section \ref{sec:pure}) and propose an algorithm that measures state novelty for any choice of action given an arbitrary density model. The key contribution of their study is drawing a connection, called \textit{pseudo-count}, between intrinsic motivation and count-based exploration. Pseudo-count quantity is derived from an arbitrary density model over the state space. The density model proposed in \cite{bellemare2016unifying} models a marginal distribution in which states are independently distributed. For any given choice of density model $ \rho $, the paper draws a connection between two unknowns: 1) pseudo-count function and 2) pseudo-count total.  The paper also introduces a connection between the conditional probability assigned to state $s$ using $ \rho $ after observing its new occurrence conditioning based on its prior observations, pseudo-count function and pseudo-count total.  The notion of information gain as a popular measure of novelty and curiosity is then shown to be related to pseudo-count, which leads to the main theorem in the paper that suggests using pseudo-count bonus leads to an exploratory behaviour as good as when the information gain bonus is used. Under two major assumptions: 1) the given density model asymptotically behaves similarly to the limiting empirical distribution, and 2) the learning rate at which $\rho$ changes with respect to the true state distribution $\mu$ is positive in the asymptotic sense, the limit of ratios of pseudo-counts to empirical counts is finite and exists for all states. \cite{bellemare2016unifying} test their proposed method in comparison with two state-of-the-art RL algorithms, Double DQN \citep{van2016deep} and A3C (Asynchronous Advantage Actor-Critic) \citep{mnih2016asynchronous} on some of the Atari 2600 games. 

In a subsequent work, \cite{ostrovski2017count} answer two questions regarding the modeling assumptions raised in \cite{bellemare2016unifying}: 1) what is the impact of the quality of density model on exploration? 2) To what extent do Monte-Carlo updates influence exploration? To address the first question, \cite{ostrovski2017count} adopt an advanced neural density model PixelCNN \citep{van2016conditional}, and discuss the challenges involved in this approach in terms of model choice, model training and model use. PixelCNN is a convolutional neural network that models a probability distribution over pixels conditioned on the previous pixels. The paper provides a list of properties that the density model requires and subsequently suggests a suitable notion of pseudo-count for DQN agents that leads to state-of-the-art results in difficult Atari games like Montezuma's Revenge.

Following the pseudo-count technique proposed by \cite{bellemare2016unifying} and \cite{ ostrovski2017count}, \cite{martin2017count} propose a new density model to measure the similarity between states and, consequently, a generalized visitation count method. Even though \cite{bellemare2016unifying} construct the density model over raw state visitations, the method proposed by \cite{martin2017count} relies on the feature map used in value function approximation to construct the density model. The bonus-based exploration algorithm $\phi$\textit{-Exploration Bonus} proposed by \cite{martin2017count} augments the extrinsic reward with the bonus term proportional to the inverse of the square root of the pseudo-count calculated based on the proposed feature-based density model. Empirically, $\phi$\textit{-Exploration Bonus} algorithm is evaluated against the $\epsilon$-greedy, A3C \citep{mnih2016asynchronous}, Double DQN \citep{hasselt2016deep}, Double DQN with pseudo-count \citep{bellemare2016unifying}, TRPO \citep{schulman2015trust}, Gorila \citep{nair2015massively}, and Dueling Network \citep{wang2016dueling} baselines in five different games from the Arcade Learning Environment (ALE).

\cite{fu2017ex2} introduce another study that uses count-based bonuses to conduct exploration in high-dimensional domains using the notions of curiosity and novelty. Effective exploration methods that are based on a notion of visitation novelty typically require either a tabular representation of states and actions or a generative model over state and actions, which can be difficult to train in high-dimensional and continuous settings. \cite{fu2017ex2} propose an approach to approximate state visitation densities using a discriminative model (exemplar model) over the complex model of states using deep neural networks, where the classifier assigns reward bonuses if the recently visited state is novel. The authors of this work show that discriminative modeling is equivalent to implicit density estimation. They argue that learning a discriminative model using standard convolutional classifier networks in the case of rich sensory inputs like images is typically easier than learning the generative model of the environment. Their proposed model is inspired by the concept of Generative Adversarial Networks \citep{goodfellow2014generative} and employs the intuition that novel states are typically more easily distinguished from all other visited states. The main idea is to maintain a density estimator using exemplar models based on a discriminatively trained classifier instead of maintaining explicit counts. To train the proposed discriminator, a cross-entropy loss is employed. \citet{fu2017ex2} evaluate their proposed method in sparse-reward continuous high-dimensional control tasks in MuJoCo \citep{todorov2012mujoco} and Vizdoom \citep{kempka2016vizdoom}. They compare the algorithm's performance with the two state-of-the-art baseline by \cite{houthooft2016vime} (discussed in Section \ref{sec:pred-bounus-fa}) and \cite{schulman2015trust}.

As an extension of count-based exploration to high-dimensional and continuous deep RL benchmarks, \cite{tang2017exploration} use a hashing mechanism to map novel states and visited states to hash codes and subsequently count the state visitations using the corresponding hash table. In the simple domains, authors propose a static hashing approach, in which the state space is discretized using a hash function such as SimHash \citep{charikar2002similarity}, and subsequently the bonus term is set to be proportional to the inverse of the square root of state count with respect to the hash code. In environments with complex structures, the authors adopt the Learned Hashing mechanism that implements an autoencoder to learn the appropriate hash codes. Like the static hash mechanism, the Learned Hashing mechanism also employs a bonus term proportional to the inverse of the square root of count on the hash codes. This approach outperforms the method presented by \cite{houthooft2016vime}  in some  rllab benchmark tasks, as well as the vanilla DQN agent in some Atari 2600 games.

Inspired by the results from \cite{wu2018laplacian} and \cite{machado2017laplacian}, authors of \cite{machado2020count} show that the inverse of $l_1$ norm of successor representation \citep{dayan1993improving} can be interpreted as an exploration bonus in both tabular and function approximation setting. Successor representation \citep{dayan1993improving} can be interpreted as an implicit estimator of the transition dynamics of the environment. In the tabular setting, they augment the Sarsa \citep{sutton1992reinforcement} update rule with the inverse of the norm of the successor representation of the visited states. Their proposed algorithm is empirically compared with traditional Sarsa in traditional PAC-MDP domains SixArms and RiverSwim \citep{strehl2008analysis}. In the function approximation case, the bonus used is similar to the one utilized in tabular setting and is the inverse of $l_1$ norm of the parameterized successor feature vector. \cite{machado2020count} evaluate their proposed algorithm empirically in the Arcade Learning Environments \citep{bellemare2013arcade} with sparse reward structure, including Freeway, Gravitar, Montezuma's Revenge, Private Eye, Solaris, and Venture.

\subsection{Prediction error-based bonus}

In this category of exploration methods, the bonus term is computed based on the change in the agent's knowledge about the environment dynamics. The agent's knowledge about the environment is often measured through a prediction model of environment dynamics. This section focuses on the exploration techniques that use Prediction Error (PE) as an exploration bonus. PE is a term used to measure the difference between the true environment model parameters and their estimates that are used to predict transition dynamics. The methods that fall into this category use the discrepancy between the induced prediction from the learned model of environment dynamics and state-action representation models, and the real observation to assess the novelty of the visited states. The states that lead to more considerable discrepancy are considered more informative than those with a smaller discrepancy. The first two studies that employ PE as an exploration bonus to encourage curiosity are \cite{schmidhuber1991curious} and \cite{schmidhuber1991possibility}, which are explained in detail in Section \ref{sec:pure} (due to the fact that they can also function in environments that do not provide extrinsic rewards).

Formally, let $H_t$ be the history of observations until time-step $t$, $a_t$ denote the action taken at time $t$, and $M_\phi$ be the predictive model of transition parameterized by the feature function $\phi$. Then, the prediction error at time $t$ is proportional to, 
\begin{align}
    e(H_{t-1},a_t,s_{t}) \propto \norm{s_{t} - M_\phi(H_{t-1},a_t)}_p,
\end{align}
where $\norm{.}_p$ denotes the p-norm of a given vector.

\begin{table}
\centering
\begin{tabular}{p{5cm}|p{5cm}|p{4cm}}
    Reference & Approach & Performance measure \\
    \hline
    &\textbf{Tabular Methods}& \\
    \hline
    \cite{schmidhuber1991curious} & Confidence-based  & Empirical-Grid World \\
    \cite{dearden1998bayesian} & Information gain-based  & Convergence/Empirical\\
    \cite{wiering1999explorations} & Confidence-based & PAC bound \\
    \cite{ishii2002control} & Confidence-based/Entropy-based & Empirical-Grid\\
    \cite{strehl2004exploration} & Confidence-based &  Empirical \\
    \cite{strehl2008analysis} & Confidence-based &  PAC bound/Empirical\\
    \cite{lopes2012exploration} & density-based & PAC bound/Empirical\\
    \hline
    &\textbf{Function Approximation Methods}& \\
    \hline
    \cite{white2010interval} & Confidence based & Convergence/Empirical\\
    \cite{stadie2015incentivizing} & PE-based bonus & Empirical\\
    \cite{houthooft2016vime} & Information gain & Empirical\\
    \cite{pathak2017curiosity} & PE-based bonus & Empirical\\
    \cite{burda2018large} & PE-based bonus & Empirical\\
    \cite{hong2018diversity} & PE-based bonus & Empirical\\ 
    \cite{burda2018exploration} & density-based & Empirical\\
    \cite{kim2019curiosity} & Information gain-based & Empirical\\
\end{tabular}
\caption{Prediction Error-based methods}
\label{tab:prediction-error}
\end{table}
 
\subsubsection{Prediction error-based bonus: Tabular}

\cite{dayan1996exploration} consider the problem of exploration in a non-stationary absorbing finite POMDP setting and provide a systematic approach to designing exploration bonuses in such setting. Their algorithm borrows the certainty equivalence approximation technique from the dual control literature and provides a statistical model of the environment's uncertainty in a finite state space setting and subsequently incorporates the uncertainty estimates into the systematic design of exploration bonuses. The exploration bonus in \cite{dayan1996exploration} is proportional to the amount of uncertainty induced by the agent's belief system and the true model of the environment. \cite{dayan1996exploration} assess their proposed method against DAYNA \citep{sutton1991dyna} in a two-dimensional maze with movable barriers. 

The concept of Interval Estimation (IE) was first introduced by \cite{kaelbling1993learning} in the bandit setting to employ confidence intervals during the exploration phase. The agent chooses the action that induces the highest upper confidence bound. A few years later, \cite{wiering1999explorations} provided a theoretical extension to \cite{kaelbling1993learning} and discussed a new variation of Model-Based Interval Estimation (MBIE) by augmenting the Bellman optimally equation with a bonus term proportional to ${1}/{\sqrt{N(s,a)}}$. The author also provide a formal PAC-style guarantee for their proposed algorithm and theoretically analyze the effect of additive bonus term on the number of time steps required to achieve the sub-optimal performance. 

In the stream of model-based approaches to exploration, \cite{ishii2002control} compute the exploration bonus using the entropy of the posterior distribution of the state-transition kernel. The reward associated with the state-action pair is composed of the obtained immediate reward, and the entropy of the visited state-action pairs. The action sampling policy is based on a softmax action selection algorithm combined with an entropic bonus term. Small entropy means that the amount of information acquired given the agent's current model of the environment is expected to be small, and therefore, the probability of taking action given the current state of the agent is small. \cite{ishii2002control} assess their proposed method in a small 2D maze environment with fixed and moving barriers and a zig-zag maze. In both experiments, the effect of exploration bonus on the required number of steps for reaching the shortest path is shown.

\cite{lopes2012exploration} improve the traditional model-based exploration techniques based on OFU principle such as R-Max \citep{brafman2002r} and Bayesian exploration bonus \citep{kolter2009near}, and empirically estimate the learner's accuracy and the learning progress. \cite{lopes2012exploration} use the change in the loss of prediction error (both mean and variance) as the bonus term, and subsequently use it to modify R-Max and Bayesian exploration bonus methods. This approach is particularly useful in scenarios where the agent has an incorrect prior knowledge of the transition model, or when it changes over time.  The learning progress is measured with respect to the empirical estimate of predictive error using the leave-one-out cross-validation estimator.

\subsubsection{Prediction error-based bonus: Function Approximation} \label{sec:pred-bounus-fa}

In this section, we focus on the exploration techniques that use PE as a measure of uncertainty to design exploration bonuses in domains with large or infinite state spaces, where methods that focus on tabular settings fail to generalize. 

\cite{stadie2015incentivizing} propose a method of exploration that hinges upon a model of system dynamics trained using past experiences and observations.  A state is considered novel and accordingly receives an exploration bonus based on its disagreement with the environment's learned model. Formally, given the state encoding function $\sigma$, the prediction error with respect to $\sigma$ at time $t$ is denoted by $e_t(\sigma)$ and thus the bonus term is proportional to ${e_t(\sigma)}/{t}$. The authors benchmark their proposed algorithm on Atari tasks and show that it is an efficient method of assigning exploration bonuses for large and complex RL domains. The predictive model introduced by \citet{stadie2015incentivizing} is a simple two-layer neural network, and the prediction error is measured with respect to the $l_2$ Euclidean norm. They evaluate their proposed method in 14 Arcade Learning Environments (ALE) against Boltzmann, DQN and Thompson sampling methods.

In another diversity-driven method, \cite{hong2018diversity} augment a diversity-based bonus with the loss function and encourage the agent to explore diverse behaviours during the training phase. The modified loss function is computed by subtracting the current policy's expected deviation or distance from a set of recently adopted policies. They use a clipping threshold in the case of observing extraordinary deviation in the computed empirical expectation. The authors evaluate the performanc of their proposed method in Mujoco and Atari environments against vanilla DDPG and the Parameter-Noise exploration method \citep{plappert2018parameter}.

\cite{burda2018large} conducted a large set of experiments on curiosity-driven learning algorithms that work with intrinsic reward mechanisms across 54 different environments. Interestingly, the results presented show the impact of feature learning on better generalizability while using prediction error as a deriving force for exploration. Through the conducted experiments, they also demonstrate the limitations of prediction-based bonus mechanisms.   

Some studies measure the observed state's novelty based on the amount of PE the observed state induces on the network that is trained using the agent's past experience. For example, \cite{burda2018exploration} introduce a simple notion of exploration bonus, which is based on the PE induced by the features of observed states and the prediction of the randomly initialized network when the environment is stochastic. \cite{burda2018exploration} count four factors as the primary sources of error in prediction, 1) the amount of training data, 2) stochasticity of environment, 3) model misspecification, and 4) learning dynamics. The uncertainty factor considered in their study is based upon the uncertainty quantification method proposed initially by \cite{osband2018randomized}. \cite{burda2018exploration} assess their proposed exploration method in the difficult Atari game Montezuma's Revenge and outperforms the state-of-the-art baselines. 

A line of research uses information gain (IG) as an exploration bonus (For a more detailed explanation regarding information gain, refer to section \ref{sec:intrinsic}). For instance, \cite{houthooft2016vime} propose a curiosity-driven strategy, which uses information gain as a driving force to encourage exploration of actions that lead to states that cause a larger change in the agent’s internal model of environment dynamics. The state and action spaces are considered to be continuous. The paper proposes a variational approach to approximate the true posterior, and therefore, the information gain is measured using the KL-divergence between the agent’s internal belief over environment dynamics at different time steps. The main challenge in their proposed model is the computation of variational lower-bound. The way \cite{houthooft2016vime} compute variational lower-bound, requires the calculation of the posterior probability, which is generally considered to be computationally intractable. The computed variational lower-bound is used to measure the agent’s curiosity. \cite{houthooft2016vime} assess their proposed algorithm in continuous Mujoco domains, and compare its performance with TRPO, ERWR and REINFORCE. 

Another study that uses IG as an exploration bonus is introduced by \cite{kim2019curiosity}. The authors apply the information bottleneck (IB) principle \citep{tishby2000information} to design an exploration bonus to handle the exploration-exploitation trade-off, particularly in distractive environments. The bonus term in Curiosity-Bottleneck (CB) objective (inspired by IB principle) appears in the form of mutual information, measured by KL-divergence between the latent representation of the environment and the input observation. To inspect the performance of the proposed CB method, \cite{kim2019curiosity} perform experiments on three environments: 1) Novelty detection on MNIST and Fashion-MNIST, 2) Treasure Hunt in a grid-world environment, and 3) Atari's Gravitar, Montezuma’s Revenge, and Solaris  games. The CB performance is compared with the work of \cite{burda2018exploration}.
\section{Deliberate Exploration}
\label{sec:prior}

The optimal solution to the exploration-exploitation problem is a strategy that yields the highest expected total reward over the entire duration of an agent's interaction with the environment. The problem of finding such an optimal strategy is a meta-problem in itself, where a notion of optimality can be defined with respect to a distribution over the models of the environments the agent is likely to encounter. Which exploration strategy works best depends on the range of environments considered plausible by the agent. If a probability distribution over the environment is known, we can define the optimal exploration strategy as the strategy that yields the highest expected total reward in expectation over this distribution. 

If unknown transition and reward model parameters are considered as the unobservable states of the system, then the entire problem can be defined as a specific kind of POMDP where the hidden part is a set of environment parameters and the observable part is the state of the sampled MDP. 
However, this formulation can have too many belief states and can thus in general not be solved exactly within practicable time. 
Various sub-fields have focused on different avenues of solving this problem approximately.  
The Bayesian approach focuses on tackling the full problem, termed Bayes-adaptive MDP, by introducing different approximations to the problem; for example by making prior assumptions about the form of the uncertainty over the unknown model parameters either via function approximation or representing the distribution over environment using a set of samples. 
On the other hand, meta-learning approaches assume the agent does not directly have access to a distribution over environments but can be trained on samples from the relevant distribution. Various methods, e.g. from model-free reinforcement learning, can then be used to find policies that can effectively embed interaction histories and map them to actions to be taken. These methods tend to aim at finding (locally) optimal solutions to the Bayes-adaptive MDP within a parametrized family of policies.

\subsection{Solving the exploration-exploitation trade-off optimally}
\label{sec:optimally-solving-exploration}

In the Bayesian approach for RL, a posterior is maintained over the possible models of the environment given the observations so far.  As the agent collects more observations, the belief is updated to reflect this new information.  
Consequently, learning bayes-optimally in an MDP is equivalent to solving for an optimal action selection strategy in a meta-level Markov decision process defined by these belief states. 
In the following sections, we will focus on a particular formulation of this meta-level problem: the 
Bayes-Adaptive MDP \citep{duff2002optimal}  formulation, where the belief state is given by a current base-level state as well as a posterior distribution over the base-level transition and reward models.

When the dynamics are unknown, the Bayesian RL formulation assumes that the transition model $\p$ is a latent variable according to some prior distribution $P(\p)$. Let $H_t$ denote the history of observations up to time $t$, then the dynamics model is updated according to the Bayes rule $P(\p|H_t) \propto P(H_t|\p) P(\p)$, in response to the observed transitions. 
The uncertainty of the model dynamics is handled by transforming it into uncertainty into the augmented state and history space. 
The actual state of the agent, $\s$, together with the belief state that that consists of parameters defining uncertainty distributions over the transition model, $\x$, comprise the \textit{hyperstate}, $\s^{+} = \s \times \x$.
The new dynamics model in this new augmented space is defined by 
$\p^{+}(s',x', a,s,x)$, that denotes the transition model for hyperstates, conditioned on action $a$ being taken in hyperstate $\langle s, x \rangle$. The reward function for the aurgmented MDP is given by $\r^{+}(s,x, a) = \r(s,a)$.
The new MDP is defined by the tuple of $M^{+} = \langle \s^{+}, \a, \p^{+}, \r^{+}, \gamma \rangle$ and is known as the \textbf{Bayes-Adaptive MDP (BAMDP)} \citep{duff2002optimal}.

The hyperstate is a sufficient statistics for the process evolving under uncertainty, and the Bellman equations formalism used for MDPs also holds true for the generalized hyperstate MDP. The solution to the Bellman equations gives the value function of the hyperstate, and an optimal value function implicitly defines the optimal learning policy, which maps hyperstates to actions. The value function given by Bellman equation for the BAMDP is given by: 
\begin{align}
    \label{eq:BAMDP-optimal}
    V^{\star}(s,x) &= \max_{a \in \a} \left[ \r^{+}(x,a) + \gamma \sum_{\substack{x' \in \x,\\ s' \in \s}}\p^{+}(s',x' \mid s,x,a)  V^{\star}(s',x') \right].
\end{align}

The agent starts in the belief state corresponding to its prior and, by executing the greedy policy in the BAMDP while updating its posterior, acts optimally (with respect to its beliefs) in the original MDP. The Bayes-optimal policy for the unknown environment is the optimal policy of the BAMDP, thereby providing an elegant solution to the exploration-exploitation trade-off. 
Through this framework, rich prior knowledge about the environment can be naturally incorporated into the planning process, potentially leading to more efficient exploration and exploitation of the uncertain world.


\begin{figure}
    \centering
    \begin{floatrow}
    \subfloat[]{
    \begin{tikzpicture}[shorten >=1pt,node distance=2.5cm,>=stealth',on grid,auto]

    \tikzstyle{every state}=[semithick,fill={rgb:black,1;white,10;green,0}]

    \node[state](s0){$-1$};
    \node[state,right of=s0](s1){$+1$} ;

    \path[->]
    (s0)edge[bend left]  node{$p_{12}^{1}$}(s1)
    (s0)edge[loop left]  node{$p_{11}^{1}$}()
    (s1)edge[bend left] node{$p_{21}^{1}$}(s0)
    (s1)edge[loop right] node{$p_{22}^{1}$}();
    \end{tikzpicture}
    }

    \subfloat[]{
    \begin{tikzpicture}[shorten >=1pt,node distance=2cm,>=stealth',on grid,auto]
    
    \tikzstyle{every state}=[semithick,fill={rgb:black,1;white,10;green,0}]
    
    \node[state](s0){$-1$};
    \node[state,right of=s0](s1){$+1$} ;
    
    \path[->]
    (s0)edge[bend left]  node{$p_{12}^{2}$}(s1)
    (s0)edge[loop left]  node{$p_{11}^{2}$}()
    (s1)edge[bend left] node{$p_{21}^{2}$}(s0)
    (s1)edge[loop right] node{$p_{22}^{2}$}();
    \end{tikzpicture}
    }
    \end{floatrow}
   
   \caption{A 2-state MDP with uncertain transition probabilities under (a) action 1 and (b) action 2. Rewards are denoted by $\pm 1$ in the states.}
   \label{fig:duff-2-state-mdp}

\end{figure}
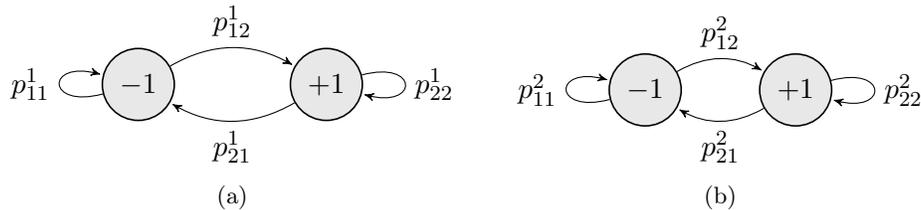

\paragraph{Example:} We use the following example from \cite{duff2003design} to highlight how the Bayesian formulation allows to solve the exploration-exploitation optimally. Consider an MDP with 2 states and 2 actions as shown in Figure~\ref{fig:duff-2-state-mdp}. The hyperstate in this case is defined by a tuple $(s;x)$, where the physical state $s$ is the state the agent currently is in as given by the MDP, and the information state $x$, which is the collection of distributions describing uncertainty in the transition probabilities. The rewards ($\pm 1$) are deterministic and are received when the agent lands in the corresponding state. The unknown transition probabilities are denoted by the labelled arcs. For this particular example, the authors propose using an appropriate conjugate family of distributions, such as Dirichlet, to model the uncertainty. For instance, if the uncertainty in $p_{11}^1$ (transition from state 1 under action 1) is represented as beta distribution parameterized by $(\alpha_1^1, \beta_1^1)$, then hyper state for $s=1$ can be written as: $(1; \begin{bmatrix} \alpha_1^1& \beta_1^1\\\beta_2^1& \alpha_2^1 \end{bmatrix}, \begin{bmatrix} \alpha_1^2& \beta_1^2\\\beta_2^2& \alpha_2^2 \end{bmatrix})$.
The new augmented transition function can be written as 
\begin{equation*}
    P(s',x' |s,x,a) = P(s'|s,x,a) P(x'|s,x,a,s').    
\end{equation*}

The first factor can be estimated from samples by taking expectation over the possible transition function to corresponding to a given hyperstate. Therefore, for a transition from $s$ under action $a$, the corresponding terms for $s'=1$ and $s'=2$ will be $\frac{\alpha^{a}}{\alpha_{s}^{a} + \beta_{s}^{a}}$  and $\frac{\beta^{a}}{\alpha_{s}^{a} + \beta_{s}^{a}}$ respectively.
For the second factor, the parameter of the dirichlet is given by  the number of 'effective' transitions of each type observed in transit from the initial hyperstate to the given hyperstate. 
For this particular example, the form of the posterior update for the information state parameters, given an observation is also a beta distribution, but with parameters that are incremented to reflect the observed data. As such, for transition from $s$ under action $a$, the update information state terms for $s'=1$ and $s'=2$ will be $\alpha_{s}^{a}+1$ and $\beta_{s}^{a}+1$. An optimality equation can be written for the local transitions and the corresponding successor hyperstates. For the above example, the optimal value function has the form:
\begin{align*}
    V\left(1; \begin{bmatrix} \alpha_1^1& \beta_1^1\\\beta_2^1& \alpha_2^1 \end{bmatrix}, \begin{bmatrix} \alpha_1^2& \beta_1^2\\\beta_2^2& \alpha_2^2 \end{bmatrix}\right) &= 
    \max \Bigg\{ 
    \frac{\alpha_1^1}{\alpha_1^1 + \beta_1^1} \left[ r_{11}^1 + V\left(1; \begin{bmatrix} \alpha_1^1 + 1& \beta_1^1\\\beta_2^1& \alpha_2^1 \end{bmatrix}, \begin{bmatrix} \alpha_1^2& \beta_1^2\\\beta_2^2& \alpha_2^2 \end{bmatrix}\right) \right]  \\
    &\quad\quad + \frac{\beta_1^1}{\alpha_1^1 + \beta_1^1} \left[r_{12}^1 + V\left(2; \begin{bmatrix} \alpha_1^1 & \beta_1^1 + 1\\\beta_2^1& \alpha_2^1 \end{bmatrix}, \begin{bmatrix} \alpha_1^2& \beta_1^2\\\beta_2^2& \alpha_2^2 \end{bmatrix}\right) \right] , \\ 
    &\quad\frac{\alpha_1^2}{\alpha_1^2 + \beta_1^2} \left[ r_{11}^2 + V\left(1; \begin{bmatrix} \alpha_1^1 & \beta_1^1 \\\beta_2^1& \alpha_2^1 \end{bmatrix}, \begin{bmatrix} \alpha_1^2 + 1& \beta_1^2\\\beta_2^2& \alpha_2^2 \end{bmatrix}\right) \right]  \\
    &\quad\quad + \frac{\beta_1^2}{\alpha_1^2 + \beta_1^2} \left[r_{12}^2 + V\left(2; \begin{bmatrix} \alpha_1^1 & \beta_1^1 \\\beta_2^1& \alpha_2^1 \end{bmatrix}, \begin{bmatrix} \alpha_1^2& \beta_1^2 + 1\\\beta_2^2& \alpha_2^2 \end{bmatrix}\right) \right] 
    \Bigg\} .
\end{align*}

An optimal policy can be computed using the dynamic programming on the augmented MDP. However, as each transition can increment any of the information state parameters with every time step, there is an exponential increase in number of distinct reachable hyperstates with the time horizon ($4^{depth}$ hyperstates at a given depth for the above example). 

Hence, in the Bayesian RL formulation the exploration-exploitation trade-off is handled in a principled manner, because the agent does not heuristically choose between exploiting or exploring, rather, it takes an optimal action (that might lead to a mixture of exploration and exploitation) with respect to its full Bayesian model of the uncertain sequential decision process \citep{duff2002optimal, poupart2006analytic}. In Bayesian decision theory, the optimal action is the one that, over the entire time horizon considered, maximizes the expected reward, averaged over the possible world models. Any gain in reducing the uncertainty over the posterior transition models is not valued just for its own sake, but rather is driven by the potential gain in the future reward that it offers.

The major problem with the BAMDPs is that the number of hyperstates grows exponentially with the time-horizon. 
This exponential growth in hyperstates limits the scalability of this approach as it can make solving the problem intractable when either the size of state and action spaces increases or the planning horizon increases.
In the next section we will go over some of the works that provide tractable computational procedures that retain the Bayesian formulation but sidesteps the intractability by employing various approximation and sampling techniques.

\subsection{Bayesian Methods} 
\label{subsec:approx bayes adaptive}

To recap, in the Bayes-Adaptive setting, a prior over MDPs is given to the agent, and (approximately)  optimal  decisions  are  made  based  on  the  posterior  over  MDPs, where the posterior is a function of the interaction history with the MDP. 
In this section we review a few notable works that are based on the different BAMDP formulations, to illustrate how the explosion of the hyperstate space is handled to provide tractable approximate solutions.
For a thorough insight into different techniques for tractability in Bayesian methods for RL we refer the reader to the survey by \citet[Chapter 4]{ghavamzadeh2015bayesian}.

\textit{Bayesian Q-Learning} \citep{dearden1998bayesian} adopts a Bayesian approach to the Q-learning by maintaining and propagating the probability distributions to represent the uncertainty over the agent's estimates of Q-values. 
Under certain modeling assumptions, the authors show that Dirichlet distributions can be used to maintain such a posterior over the Q-values.
Instead of solving the BAMDP using dynamic programming as in the previous Section~\ref{sec:optimally-solving-exploration}, the authors propose an action selection procedure based on the `value of information' - the expected gain in future decision quality that might arise from information acquired from the current action choice. Intuitively, this notion considers the gain that can be acquired learning the true value of a particular Q-value. 
Formally, let $a_1$ and $a_2$ denote the actions with best and second best expected values respectively, and $q^{\star}_{s,a}$ denote a random variable representing the a possible value of $Q^{\star}(s,a)$ in some MDP, then the gain from learning the true value is defined as:
\begin{equation}
    \label{eq:BQL-gain}
    Gain_{s,a}(q^{\star}_{s,a}) = \begin{cases}
        \E[q(s,a_2)] - q^{\star}_{s,a} & \text{if $a=a_1$ and $q^{\star}_{s,a} < \E[q_{s,a_2}]$}, \\
         q^{\star}_{s,a} - \E[q(s,a_1)] & \text{if $a \neq a_1$ and $q^{\star}_{s,a} > \E[q_{s,a_1}]$}, \\
        0 & \text{otherwise}
    \end{cases}
\end{equation}
As the agent doesn't know the true value of $q^{\star}_{s,a}$, the expected gain is computed using the prior beliefs to estimate the Value of Perfect Information (VPI):
\begin{equation}
    \label{eq:VPI}
    \textit{VPI}(s,a) = \int_{- \infty}^{+ \infty} Gain_{s,a}(x) Pr(q_{s,a}=x) dx,
\end{equation}
The value of perfect information gives an upper bound on the myopic (1-step) expected value of information for exploring with action $a$. In order to take into account the exploitation aspect, the expected reward is also added to the action selection criteria. Therefore, the goal is to select action that maximizes $( \E[q_{s,a}] + \textit{VPI}(s,a) )$.

Once the action is taken and transitions are observed, the authors propose two ways of estimating the distribution of the Q-value. The first one is based on Moment matching that leads to a closed-form update but can become overly confident. The second technique is based on mixture updates that are more cautious but require numerical integration. Finally, they provide some theoretical results on the convergence of the algorithm and then they conclude with some experimental results on three toy problems (a 5-state chain MDP, an 8-state loop MDP and a 2D-maze of size 8x8) and compare their work with three other methods.

\begin{table}
    \centering
    \begin{tabularx}{\columnwidth}{c|c|c}
        Approach & Choice of approximation & Solution method \\
        \hline
        \cite{dearden1998bayesian} & Online myopic value function  & Dynamic programming 
        
        \\
        \cite{duff2002optimal} & Offline value function  & Policy gradient 
        \\
        \cite{wang2005bayesian} & Online tree search  & Tree backups, myopic \\
        & &  heuristic at leaves  
        \\
        \cite{poupart2006analytic} & Offline value function  & Point-based POMDP methods 
        \\
        \cite{guez2012efficient} & Online tree search  & Q-learning with rollout policy 
        \\
    \end{tabularx}
    \caption{Overview of the main techniques covered in Section~\ref{subsec:approx bayes adaptive}}
    \label{tab:bayesian-methods}
\end{table}

\textit{Optimal Probe} \citep{duff2003design} retains the full Bayesian framework but proposes to sidestep the intractable calculations by using a novel actor-critic architecture and proposing a corresponding policy-gradient based update for it.
The policies and value functions are approximated by functions involving linear combinations of the information state components. 
The main assumption is that the value function is a relatively smooth function of the information state, $x$.  
For the feature set, they propose using the components of the information state. As such, the critic is parameterized as: $V(s,x) \approx V_{s}(x) = \sum_l \theta_l [s] x_l,$ where  $V_s$ represents the function approximator associated with the state in the original MDP $s$, and $\theta_l$ represents the parameters corresponding to the $l$-th information state component. 
For policies, they propose using a separate parameterized function approximator for each original state $s$ and possible action. This is because the stochastic policies that map from hyperstates to actions are required to be relatively smooth over the information state $x$, but should allow arbitrary and discontinuous variation with respect to the original state and the prospective action.
This reduces the size of the class of stochastic policies that the actor can model, but the hope is that this parameterized family of policies will be rich enough to represent \textit{near-optimal} policies. 

For updating the policy, a Monte-Carlo based Policy gradient update rule is proposed that uses a single hyperstate trajectory for providing an unbiased estimate of the gradient components with respect to the actor's parameters. In conclusion, the assumed function class for the actor introduces a bias but makes the complexity independent of the (exponential) number of hyperstates. Policy gradients can be expressed in terms of a matrix representing the steady-state probability of hyperstates. This matrix is again computationally intractable but can be approximated using sampled roll-outs. They test their approach a 2-states and 2 actions toy MDP with a horizon of 25 and were the first to show a tractable algorithm for that case. This method however is only computationally feasible for domains with a small number of information states where the proposed architecture works.

\textit{Sparse sampling} \citep{kearns2002sparse} is a sample-based tree search algorithm, where the agent samples the next possible tree nodes from each state and then applies Bellman backup to propagate the values of child nodes to the parent node. In \textit{Bayesian Sparse Sampling} \citep{wang2005bayesian} the authors apply the sparse sampling technique to search over BAMDPs, where the task is to use lookahead search to estimate the long term value of possible actions in a given belief state. 
The key idea is to exploit information in the Bayesian posterior to make intelligent action selection decisions during the look-ahead simulation, rather than simply enumerating over all the actions or selecting the actions myopically.  The search tree is expanded adaptively in a non-uniform manner, instead of building a uniformly balanced look-ahead tree. The intuition is that the agent only needs to investigate actions that are potentially optimal, and in this way can save computation resources on sub-optimal actions.
At each decision node, a promising action is selected using a heuristic based on Thompson sampling \citep{thompson1933likelihood} to preferentially expand the tree below actions that appear to be locally promising. At each branch node, a successor belief-state is sampled from the transition dynamics of the belief-state MDP. Once chosen, the action is executed, and a new belief state is entered. As the focus of the work is to demonstrate action selection improvements, they compare their approach with other selection schemes like standard Sparse sampling, Thompson sampling, Interval Estimation, and Boltzmann exploration techniques in the continuous 2-dimensional action space Gaussian processes task.  The results show that their approach yields improved action selection quality whenever Bayesian posteriors can be conveniently calculated.

\cite{poupart2006analytic}, focus on the problem of discrete Bayesian model-based RL in the online setting, and propose the \textit{BEETLE} (Bayesian Exploration Exploitation Tradeoff in LEarning), a point-based value iteration algorithm, that takes into account exploration during the exploitation step itself using the belief states mechanism.  
The main contribution of the work is that they present an analytical derivation of the optimal value functions for the discrete Bayesian RL problem where the optimal value function is parameterized by a set of multivariate polynomials. 
This analytical form then allows them to build an efficient point-based value iteration algorithm that exploits the particular form of parameterization.  As a result, they have a computationally efficient offline policy optimization technique and results that show the optimization remains efficient as long as the number of unknown transition dynamics parameters remains small.  The results are based on the argument that the transition dynamics of many problems can be encoded with few parameters by either tying the parameters together or using a factored model. The algorithm achieves online efficiency, by moving the policy optimization offline and doing only action selection and belief monitoring at run time. The authors compare their proposed method with two-heuristics based exploration approaches on two discrete toy MDPs benchmarks: a toy-chain with 5 states and 2 actions and an assistive technology scenario MDP with 9 stats and 6 actions.


The  \textit{Bayes-Adaptive Monte Carlo Planning} (BAMCP) algorithm \cite{guez2012efficient} provides a sample-based method for approximate Bayes-optimal planning for discrete MDPs that exploits Monte-Carlo tree search (MCTS). The core idea is to use the UCT algorithm \citep{kocsis2006bandit} in a computationally efficient manner for BAMDPs, where the belief state is approximated using the samples sampled only at the root node of the tree. 
The authors propose Bayes-Adaptive UCT, where instead of integration over all the transition models, or even approximating this expectation using an average of sampled transition model, only a single transition model $\p^i$ is sampled from the agent's current belief (posterior at the root of the search tree) and is used to simulate all the necessary samples during this episode.  A tree policy then treats the forward search as a meta-exploration problem, in a similar manner as the vanilla UCT problem. The goal of the tree policy is to exploit regions of the search tree that appear better than the others while continuing to explore the less known parts of the tree. For the exploitation part, a rollout policy is learned in a model-free manner, using Q-learning, from the samples collected by the agent as a result of the interaction with the environment.
In order for further computational efficiency, the authors propose a novel lazy sampling scheme for the partial transition models. The intuition is that if the transition parameters for different states and actions are independent, then instead of sampling a complete $\p$, only the parameters necessary for individual state-action pairs can be sampled. 
The returns from each episode are then used to update the value of each node in the search tree during the planning. By integrating over many simulations, and therefore many sampled MDPs, the optimal value of each future sequence is obtained with respect to the agent's belief.

The authors compare their method with BOSS \citep{asmuth2009bayesian} and BEB \citep{kolter2009near} in the discrete grid-world domain (10 x 10 states) and loop domain with 9 states and show their method outperforms the others. They also test their method on an infinite 2D grid-world domain where they show that their method greatly outperforms the other baselines.
In the infinite 2D grid domain the baselines can not handle the large state space but as BAMCP limits the posterior inference to the root of the search tree it is not directly affected by the size of the state space, but instead is limited by the planning time.

\subsection{Meta-learning}

Some exploration strategies are based on, or emerge from, a meta learning perspective. 
Meta learning focuses on learning an appropriate bias from a collection of tasks that allows more efficient learning on new, similar tasks \citep{vilalta2002perspective}. The field of meta-reinforcement learning uses this approach in reinforcement learning settings. As such, the agent interacts with multiple train MDPs, allowing it to learn a strategy for interacting with eventual novel test MDPs from the same family. At `meta-train time' the agent learns general patterns from a set of train tasks, that can be exploited to learn more efficiently at `meta-test time'.  Usually, the train and test tasks are assumed to be drawn from the same distribution. Thus, the agent can tailor the learning strategy employed at `meta-test time' using inductive biases extracted at `meta-train time'. As a simple example, the training MDPs might be used to optimize the learning rate used on testing MDPs. However, much more complex meta-strategies might also be learned, including the update rule itself.

Before we give a more detailed breakdown of meta-reinforcement learning methods and exploration strategies used with or emerging from those methods, it is good to realize the connection between meta-reinforcement learning and Bayes-adaptive reinforcement learning. In Bayes-adaptive learning, a prior over MDPs is known to the agent, and (approximately) optimal decisions are made based on the posterior over MDPs. The posterior is a function of the interaction history with the MDP. 
Compare this set-up with a common set-up for meta-reinforcement learning where train and test tasks are assumed to be drawn from the same distribution. An optimized mapping from the interaction history to the next action to be taken can thus be seen to target the same objective as Bayes-adaptive learning, with the prior represented by a finite set of sampled train MDPs. Thus, meta-reinforcement learning strategies have the potential to learn an approximately optimal exploration-exploitation trade-off  with respect to the distribution of training MDPs.
Whereas the Bayes-adaptive literature has typically focused on discrete MDPs and tabular representations, most work on meta reinforcement learning focuses on the function approximation case, using deep neural networks to represent policies or value functions. 

Meta-reinforcement learning techniques can be classified based on the amount of structure used in defining the mapping from interaction history to actions \citep{finn2018learning}. At the extreme, such policies can be black boxes directly mapping from interaction histories to actions. We can think of this black box as combining two functions usually implemented by separate functions: updating the policy, and executing the policy in the current state. Thus, black box meta-reinforcement learning approaches are sometimes described as learning a reinforcement learning algorithm \citep{wang2016learning,duan2016rl}. 

Other classes of meta-reinforcement learning tasks impose more structure on the mapping from interaction histories to actions. One common type of structure is that the policy update is given by gradient ascent. The agent's objective then becomes finding prior policy parameters such that a few gradient steps result in good `post-update' policies \citep{finn2017model}. Another common type of structure is when, similar to Bayes-adaptive approaches, the inference of the posterior distribution over MDPs is decoupled from the action selection mechanism \citep{rakelly2019efficient,zintgraf2019variational}. 

Per category, we will now discuss how exploration can be done.  In our review, we will focus on papers that explicitly introduce exploration methods or explicitly discuss or analyze exploration behavior in these methods. Table~\ref{tab:metamethods} provides an overview of the main methods discussed and some of their characteristics.

\begin{table}
    \centering
    \begin{tabular}{l|l|p{4 cm}}
    \emph{Black box methods} & \multicolumn{2}{l}{ \emph{Meta-learned object}}  \\
    \cite{heess2015memory}  & \multicolumn{2}{l}{Recurrent policy} \\
    \cite{duan2016rl}  & \multicolumn{2}{l}{Recurrent policy} \\    
    \cite{wang2016learning}  & \multicolumn{2}{l}{Recurrent policy} \\  
    \cite{garcia2019meta}  & \multicolumn{2}{l}{Adviser policy} \\ 
    \cite{alet2020meta}  & \multicolumn{2}{l}{Bonus mechanism} \\ 
    \hline
    \emph{Gradient-based methods} & \multicolumn{2}{l}{ \emph{Exploration characteristics}}  \\
    \cite{finn2017model} & \multicolumn{2}{l}{Early gradient-based approach} \\  
    \cite{stadie2018some}  &
    \multicolumn{2}{l}{More credit to pre-update policies} \\  
    \cite{rothfuss2019promp}  &
    \multicolumn{2}{l}{Low-variance, improved action-level credit assignment.} \\  
    \cite{frans2018meta}  & \multicolumn{2}{l}{Efficient exploration through meta-learning sub-policies} \\
    \hline
    \emph{Inference-based methods} & \emph{Inference type} & \emph{Use of posterior} \\
    \cite{gupta2018meta} & Test-time approximate inference & Posterior sampling \\
    \cite{rakelly2019efficient} & Amortized inference & Posterior sampling \\
    \cite{zintgraf2019fast} & Amortized inference & Conditioning on variational parameters \\
    \end{tabular}
    \caption{Overview of the main techniques covered in this section.}
    \label{tab:metamethods}
\end{table}

\paragraph{Black-box methods}
In black-box models, a mapping from interaction history to actions is optimized directly. The interaction history could comprise states or observations, actions, and rewards at all previous time steps, or even previous episodes in the same MDP. Since interaction histories are sequences without a fixed size, recurrent neural networks are a popular architecture to represent policies or value functions. Conceptually, these methods are quite straightforward: in theory any well-known policy search method could be used, using a recurrent architecture and with interaction histories rather than single states as input. A generic policy for a black-box method is of the functional form 
\[ \pi_{\vec{\theta}}(A_t | S_t, H_t), \quad H_t = [S_0, A_0, R_0, \ldots, S_{t-1}, A_{t-1}, R_{t-1}, S_t], \]
with $\pi_{\vec{\theta}}$ the (usually recurrent) policy architecture parametrized by $\vec{\theta}$ and $H_t$ the interaction history up to time $t$.

Early work on this approach was performed by \citet{heess2015memory}, who looked at different kinds of partial observability, including a BAMDP-like setting where the agent had to explore a tank of water for a hidden platform, and thereafter exploit this knowledge to find the platform again. Work by \citet{duan2016rl} and \citet{wang2016learning} further investigated this type of approach. These works also formalized the idea of ``learning a reinforcement learning algorithm''. Although the framework is roughly similar, the methods differ in design choices such as which base reinforcement learning method is used (A2C/A3C \citep{mnih2016asynchronous} versus TRPO \citep{schulman2015trust}). Both papers look at the exploration/exploitation trade-off in bandits and visual navigation tasks compared to classical exploration methods based on Thompson sampling or exploration bonuses. On the discrete tasks with known dynamics, the meta-learning approach performs competitively with classical methods, but it is applicable to the challenging visual exploration task where these tabular methods are not applicable. In addition, \citet{duan2016rl} investigate exploration behavior on tabular MDPs, and \citet{wang2016learning} investigate multiple tasks inspired by behavioral science and neuroscience paradigms. Their `dependent arms' bandit experiments reveal that the method successfully learns the optimal exploration/exploitation strategy for this particular family of bandits, where after one exploration action without any reward the agent switches to pure exploitation behavior. This exploration behavior is not matched by the considered classical bandit algorithms. Overall, the black-box model is very general as well as conceptually straightforward, however, training the recurrent models tends to take a lot of samples and training time. 

An interesting approach is that by \citet{garcia2019meta}. Here, an `advisor' policy is learned in a black-box fashion. Since the black box environment contains a base reinforcement learning algorithm (such as REINFORCE \citep{williams1992simple} or PPO \cite{schulman2017proximal}), the meta state becomes a combination of the current MDP, the MDP state, and the base learner's internal state. This structure is reminiscent of gradient-based methods (covered in the next paragraph), although unlike in those methods, gradients are not taken through the internal update of the base learner. Actions executed in the environment are mostly based on those of the base learner, but in a fraction $\epsilon$ of time steps an explorative action from the `advisor' (meta-policy) is executed (rather than an action chosen uniformly as in $\epsilon$-greedy strategies). The authors provide the theoretical result that solving the meta MDP indeed results in optimal exploration policies in the sense of maximizing total return over a set number of episodes averaged over the prior MDP distribution. Furthermore, they show strong empirical performance in the function approximation setting compared to the base algorithms without advisor, random exploration, and the MAML meta-learning approach \citep[covered in the next section]{finn2017model} on continuous control tasks such as the `Ant' task from the Roboschool environment\footnote{\url{https://openai.com/blog/roboschool/}}.

Where the methods covered above all meta-learned a policy, \cite{alet2020meta} proposed a different approach. Their method meta-learns  curiosity mechanisms or bonuses that generalize across very different reinforcement-learning domains. They formulate the problem of finding exploration bonuses as an outer loop that will search over a space of bonuses (meta-learned), and an inner loop that will perform standard reinforcement learning using the adapted reward signal. They propose to do the meta-learning of the bonus in the combinatorial space of programs instead of transferring neural network weights resulting in an approach similar to neural architecture search. 
The programs are represented in a domain-specific language that includes modular building blocks like neural networks that can be updated with gradient-descent mechanisms, ensembles, buffers, etc.
They show that searching through a rich space of programs yields novel designs that work better than human-designed methods such as those proposed by \cite{pathak2017curiosity, burda2018exploration}. At the same time, the proposed approach generalizes across environments with different state and action spaces, for instance, image-based 2D gridworld games and Mujoco environments like Acrobot.

\paragraph{Gradient-based methods}
Gradient-based methods aim to introduce more structure in the meta reinforcement learning problems compared to the discussed black-box methods. These black-box methods in essence learn a RL algorithm specific to the current distribution over MDPs. Thus, it should be no wonder that updates take many steps at meta-train time. Gradient-based methods introduce prior knowledge about typical reinforcement learning updates in the learning process. These methods try to learn a policy in such a way that a few updates (or even a single one) results in a `post-update' policy that is able to attain high expected returns. As a function of the interaction history, the policy is then of the form
\[\pi_{\vec{\theta}'(H_t)}(A_t|S_t), \quad \vec{\theta}'(H_t) \leftarrow \vec{\theta} + \alpha \nabla_{\vec{\theta}} \mathbb{E}\left[\sum_{u=0}^{t-1} R(S_u, A_u) \right], \]
with $\vec{\theta}'$ the post-update parameters, that result from an inner update using an estimate of the policy gradient, and $\vec{s}_u, \vec{a}_u$ with $u<t$ taken from the interaction history $\vec{h}_t$ \citep{finn2017model}. 

\citet{stadie2018some} extended earlier work on gradient-based meta-RL \citep{finn2017model} with the explicit aim to improve exploration. Where the earlier implementation of model-agnostic meta learning (MAML)  by 
\citet{finn2017model} did not properly assign credit to pre-update trajectories, the proposed algorithm was hypothesized to explore better and thus called E-MAML. The proposed approach was tested on benchmark problems including mazes and `Krazy World', an environment specifically designed to test exploration in meta learning. The proposed approach indeed learned faster on these benchmarks. Furthermore, a separate analysis looked at the difference in exploration metrics (such as the number of goal states visited) and confirmed the proposed algorithm scored higher. Similar results held for E-RL$^2$, an approach based on RL$^2$ \citep{duan2016rl}, inspired by E-MAML, which attempts to promote exploration behavior in black-box methods by ignoring rewards obtained during exploratory roll-outs. 

MAML and E-MAML were analyzed in more detail by \citet{rothfuss2019promp}. They found that the MAML formulation takes the internal structure of the policy update better into account. When all terms of the gradient of this formulation are taken into account, one should thus expect better performance. To do so, they propose a low-variance estimator of the required Hessian. Their experiments on various locomotion benchmarks confirm this performance improvement. Furthermore, they explicitly analyze exploration behavior. This qualitative analysis shows that the original MAML implementation does not learn a good exploration strategy, with E-MAML doing better but having a hard time assigning credit to individual actions. The proposed LVC estimator, on the other hand, developed good exploration behavior and was able to exploit the gleaned information. 

The approach proposed by \citet{frans2018meta} meta-learns a shared hierarchy, meaning that a common set of sub-policies is learned while a master policy that selects between the sub-policies is adapted to the meta-test task at hand. This approach can be compared to methods such as the previously discussed MAML \citep{finn2017model}, although here the parameters of the shared hierarchy are not updated in the `inner' update, and second order gradients are not passed  back to the `outer' meta-level updates of the shared hierarchy. \citet{frans2018meta} find that the proposed method can successfully learn a meta-structure where exploration takes place efficiently on the level of the master policy. Furthermore, they find that sub-policies learned on small mazes can be transferred effectively to a more challenging sparse-reward navigation task.

\paragraph{Inference-based methods}
In inference-based methods, the inference of the properties of the current MDP is decoupled from taking actions in that MDP. This structure mirrors classical strategies for solving POMDPs \citep{monahan1982state} or BAMDPs  \citep{strens2000bayesian,duff2002optimal}. Since meta-learning methods are often applied to problems with continuous states and non-linear dynamics, inference over tasks can generally not be performed in closed form.  Instead, inference-based meta-reinforcement learning algorithms tend to use a learned latent embedding space and often employ some form of approximate inference \citep{gupta2018meta, rakelly2019efficient,zintgraf2019variational}. A generic inference-based policy architecture could thus be described as
\[ \pi_{\vec{\theta}}(A_t | S_t, \vec{\phi}^*), \quad
\vec{\phi}^* = \arg \min_{\vec{\phi}} \operatorname{KL}(q_{\vec{\phi}}(Z) || p( Z| H_t) )
, \]
with $\phi^*$ the optimal parameters for a variational distribution $q$ over latent context variables $Z$.

Model agnostic exploration with structured noise (MAESN), finds an approximate posterior using gradient ascent at meta-test time \citep{gupta2018meta}. Latent context variables can then be drawn in a posterior-sampling like manner. 
 These latent variables are constant during an episode and thus allow structured exploration behavior. 
The  experience with meta-train tasks is thus used both to initialize a policy and to acquire a latent exploration space that can inject structured stochasticity into a policy. The method is compared experimentally to MAML, RL$^2$, and conventional (non-meta) learning strategies. The authors find that MAESN strongly outperforms baseline strategies in terms of learning speed and performance after 100 learning iterations. Furthermore, qualitative analysis shows MAESN is able to learn a well-structured latent space that effectively explores in the space of coherent strategies for the trained family of environments. 

Both other methods perform inference in this latent space by training an amortized inference network. 
 \citet{rakelly2019efficient} optimize this network to directly minimize a chosen loss function while staying close to a prior. The resulting posterior is then used for posterior sampling. The authors find improved results compared to earlier meta-learning methods, presumably thanks to the structured and efficient exploration as well as the ability to use off-policy data offered by the decoupling of inference and acting. 

The approach by \citet{zintgraf2019variational}, instead, explicitly optimizes the embedding space to decode transition and reward information. Also, instead of posterior sampling, the policy is conditioned on the mean and co-variance of the full variational posterior. In preliminary experiments, the authors show that the approximate posterior can be used to strategically and systematically explore gridworlds. In these experiments, the proposed method outperformed black-box meta-learning methods by a large margin.

\section{Probability Matching}
\label{sec:thompson}

An entire body of algorithms for efficient exploration is inspired by the Probability Matching approach, also known as Thompson Sampling \citep{thompson1933likelihood}.
Probability matching (or Thompson Sampling)  is a heuristic for balancing the  exploration-exploitation dilemma in the Multi-Arm Bandit setting \citep{li2012open, agrawal2012analysis}. In this setting, the agent maintains a posterior distribution over its beliefs regarding the optimal action, but instead of selecting the action with the highest expected return according to the belief posterior, the agent selects the action randomly according to the probability with which it deems that action to be optimal. This approach uses the variance of the posterior to induce randomization and incentivizes the exploration of uncertain states and actions. As more experience is gathered, the variance of the posterior will decrease and concentrate on the true value. Thompson sampling is provably efficient for the bandit setting \citep{russo2013eluder}.

We use the setting from \cite{agrawal2012analysis} to give an example of the Thompson Sampling algorithm for Bernoulli bandit setting, i.e. when the agent gets a binary reward (0 or 1) for selecting an arm $i$, and the probability of success is $\mu_i$. The algorithm maintains Bayesian priors on the Bernoulli means $\mu_i$. The algorithm initially assumes arm $i$ to have a uniform prior on $\mu_i$ ($\texttt{Beta}(1,1)$). At time $t$, having observed $S_i(t)$ successes and $F_i(t)$ failures plays of the arm $i$, the algorithm corresponding updates distribution on $\mu_i$ as $ \texttt{Beta}(S_i(t) + 1, F_i(t) + 1)$. The algorithm then samples the model of the means from these posterior distributions of the $\mu_i$'s and plays an arm according to the probability of its mean being the largest.

\paragraph{Posterior Sampling for Reinforcement Learning (PSRL)} \textit{Bayesian dynamic programming} was first introduced in \citet{strens2000bayesian} and is more recently known as \textit{posterior sampling for reinforcement learning} (PSRL) \citep{osband2013more}. PSRL can be thought of as an extension of the Thompson Sampling algorithm to the RL setting with finite state and action spaces.  
Compared to Thompson Sampling, where a model is re-sampled at every time-step\footnote{In the bandit setting the length of an episode is 1 time-step.}, PSRL samples a single model for an episode and follows this policy for the duration of the episode.

In PSRL, the agent starts with a prior belief over the model of the MDP 
and then proceeds to update its full posterior distribution over models with the newly observed samples. For each episode, a model hypothesis is then sampled from this distribution, and traditional planning techniques are used to solve the MDP and obtain the optimal value function. For the current episode, the agent follows the greedy policy with respect to the optimal value function. They evaluate their approach on a 6-state chain MDP with 3 actions and a random MDP with 10 state and 5 actions. They show that PSRL outperforms UCRL2 by a large margin in both the above domains.

Although, both categories of the methods maintain a distribution over the rewards and transition dynamics obtained using a Bayesian modeling approach, PSRL based methods employ the posterior sampling exploration algorithm that requires solving for an optimal policy for a single MDP in each iteration. As such, PSRL is more computationally efficient compared to typical Bayesian-Adaptive algorithms that find optimal exploration strategy via either dynamic programming or tree look-ahead in the Bayesian belief state space over a set of a prior distribution over MDPs.

\textit{Best of Sampled Set} (BOSS) \citep{asmuth2009bayesian} drives exploration by sampling multiple models from the posterior and combining them to select actions optimistically.   The proposed algorithm resembles RMAX \citep{brafman2002r} in the sense that it samples multiple models from the posterior only when the number of transitions from a state-action pair exceeds a certain threshold. The sampled models are then merged into an optimistic MDP which is solved to select the best action. They show that this approach leads to sufficient exploration to guarantee finite-sample performance guarantees.  They compare their approach against BEETLE and RMAX and show superior results on the 5-state chain problem and 6x6 grid-world.

\cite{agrawal2017optimistic} propose an algorithm based on posterior sampling that achieves near-optimal worst-case regret bounds when the underlying MDP is communicating with (unknown) finite diameter. The diameter $D$ is defined as an upper bound on the time it takes to move from any state $s$ to any other $s'$ using an appropriate policy, for each pair of $s, s'$. The algorithm combines the optimism in the face of uncertainty principle (Section~\ref{Bonus}) with the posterior sampling heuristic. The algorithm proceeds in epochs, where, at the beginning of each epoch the algorithm generates $\psi = \Tilde{O}(S)$ sample transition probability vectors from a posterior distribution for every state and action. It then proceeds to solve the extended MDP with $\psi A$ actions and $S$ states formed using these samples. The optimal policy found from the extended MDP is then used for the entire epoch. This algorithm can be viewed as a combination of methodologies from BOSS and PSRL algorithms described above. The main contribution of this work is providing tighter regret bounds, and as such, they do not provide any experimental results for their algorithm.

\paragraph{Randomized Value Functions}  
The PSRL approach is limited to the finite state and action setting, where learning the model and planning might be tractable. For the rest of the section, we will look at the approaches that aren't based on modeling the MDP transition and rewards explicitly, but instead focus on estimating distributions over the value functions directly. The underlying assumption of these approaches is that approximating the posterior distribution for the value function is more statistically and computationally efficient than learning the MDP. \cite{osbandRVF, osband2017deep} proposed a family of methods called \textit{Randomized Value Functions} (RVFs) in order to improve the scalability of PSRL. At an abstract level, RVFs can be interpreted as a model-free version of PSRL. These methods directly model a distribution over the value functions instead of over MDPs. The agent then works by sampling a randomized value function at the beginning of each episode and following that for the rest of the episode.  Exploring with dithering strategies (Sec.~\ref{sec:Random},~\ref{sec:stochastic},~ \ref{sec:value-policy-search}), is inefficient as the agent may oscillate back and forth, it might not be able to discover temporally extended interesting behaviours. On the other hand, exploring with Randomized Value Functions, the agent is committed to a randomized but internally consistent strategy for the entire length of the episode. The switch to value function modelling also facilitates the use of function approximation. 

In order to scale posterior sampling approach to large MDPs with \textit{linear} function approximation,  \citet{osband2016generalization} introduce \textit{Randomized Least Square Value Iteration (RLSVI)} that involves using Bayesian linear regression for learning the value function. 
The goal is to extend PSRL to value function learning, that would involve maintaining a belief distribution over candidates for the optimal value function. Before each episode, the agent would then sample a value function from its posterior distribution and then apply the associated greedy policy throughout the episode. 

Least Square Value Iteration (LSVI) \citep{sutton1998introduction, szepesvari2010algorithms} performs a linear regression for the Bellman error at each timestep - similar to Fitted Q-Iteration  \citep{riedmiller2005neural}. As the value function learned from LSVI has no notion of uncertainty, algorithms based on just LSVI have to rely on other exploration strategies, like blind exploration (Sec.~\ref{sec:Random}). 
RLSVI also performs linear regression for one-step Bellman error but it incorporates a Gaussian uncertainty estimate for the resultant value function. This is equivalent to replacing the linear regression step of LSVI with a Bayesian linear regression as if the one-step Bellman error was sampled from a Gaussian distribution. Even though this is not the correct Bayesian distribution, \cite{osband2016generalization} show that it is still useful for approximating the uncertainty. As the RLSVI updates the value function based on a random sample from this distribution, the resultant value function is also a random sample from the approximate posterior. RLSVI is a provably efficient algorithm for exploration in large MDPs with linear value function approximators \citep{osband2017deep}.
The authors compare their approach with the dithering based strategies in a didactic chain environment where RLVSI is able to scale up to 100 state length chain. They also show better learning performance of RLVSI compared LSPI and LSVI on learning to play Tetris task, and a recommendation engine task, both of which have exponential state space but they have access to the appropriate basis functions for the task.

One of the problems with this approach is that the distributions over the value functions can be as complex to represent as distributions over transition model, and exact Bayesian inference might not be computationally tractable. RLSVI does not explicitly maintain and update belief distributions, as a coherent Bayesian method would, but still serves as a computationally tractable method for sampling value functions.

\begin{table}
    \centering
    \begin{tabularx}{\columnwidth}{p{0.25\textwidth}|p{0.4\textwidth}|p{0.35\textwidth}}
        Approach & Posterior Sampling & Theoretical properties \\
        \hline
        \cite{strens2000bayesian, osband2013more, osband2017posterior} & Sample 1 MDP model per episode  & Bounded Expected regret \newline $\Tilde{O}(HS\sqrt{AT})$, \newline Bounded Bayesian regret \newline $\Tilde{O}(H\sqrt{SAT})$
        \\
        \cite{asmuth2009bayesian} & Sample $K$ MDP models per step & PAC-MDP
        \\
        \cite{osband2016generalization} & Sample 1 value function per episode & Bounded Expected regret \newline $\Tilde{O}(\sqrt{H^3SAT})$
        \\
        \cite{agrawal2017optimistic} & Sample $\Tilde{O}(S)$ transition probability vectors per epoch &  Bounded Worst-case regret \newline $\Tilde{O}(D\sqrt{SAT})$
        \\
        \cite{osband2016deep} & Sample 1 head Q-network from the bootstrap ensemble per episode & - \\ 
        \cite{azizzadenesheli2018efficient} & Sample weights for last layer of Q-network per episode & - \\
        \cite{touati2018randomized} & Sample noise variables for the normalizing flow per episode & - \\
        \cite{janz2019successor} & Sample weights for the Q-network based on the successor features for every episode & - \\
    \end{tabularx}
    \caption{Overview of the main techniques covered in Section~\ref{sec:thompson}. For the theoretical properties column, $S$ and $A$ denote the cardinalities of the state and action spaces, $T$ denotes time elapsed, $H$ denotes the episode duration, and $D$ denotes the diameter.
    }
    \label{tab:thompson}
\end{table}

\citet{osband2016deep} propose bootstrapped Q-learning, an RVF based approach, as an extension of RLSVI to nonlinear function approximators. Bootstrapped-DQN consists of a simple non-parametric bootstrap\footnote{Bootstrap uses the empirical distribution of a sampled dataset as an estimate of the population statistic.} with random initialization to generate approximate posterior samples over Q-values. This technique helps in the scenario where exact Bayesian inference is intractable, such as in deep networks. Bootstrapped-DQN consists of a network with $K$ bootstrapped estimates of the Q-function, trained in parallel. This means that each $Q_1, \dots, Q_K$ provide a temporally extended (and consistent) estimate of the value uncertainty. At the start of each episode, the agent samples one head, which it follows for the duration of the episode. Bootstrapped-DQN is a non-parametric approach to uncertainty estimation. The authors also show that, when used with deep neural networks, the bootstrap can produce reasonable estimates of uncertainty. They compare their method against DQN \cite{mnih2015human} on the didactic chain MDP with 100 states, and on the Atari domain on the Arcade Learning Environment \citep{bellemare2013arcade}, where they show that Bootstrapped-DQN is able to learn faster and also improves the final score in most of the games.

Many other exploration methods, such as \citep{azizzadenesheli2018efficient, touati2018randomized} can be interpreted as combining the concept of RVF with neural network function approximation. It allows these methods to scale to high-dimensional problems, such as Atari domain \citep{bellemare2013arcade}, that otherwise might be too computationally extensive for PSRL. However, the approximations introduced in these works come with trade-offs that are not present in the original PSRL work.  Specifically, because a value function is defined with respect to a particular policy, constructing posterior over the value functions requires selection of a reference policy or distribution over policies.  
However, in practice, the above methods do not enforce any explicit structure or dependencies. \cite{janz2019successor} propose Successor Uncertainties (SU), a scalable and computationally cheaper (compared to Bootstrapped DQN) model-free exploration algorithm that retains the key elements of the PSRL.  SU models the posterior over rewards and state transitions directly and derives the posterior over the value functions analytically, thereby ensuring the posterior over the values estimates matches the posterior sampling policy. Empirically, SU performs much better on hard tabular didactic chain problem where the algorithm scales up to chains of length 200 states. On the Atari domain, SU outperforms the closest RVF algorithm - Bootstrapped-DQN on 36 of 49 games.

\section{Conclusion and Perspectives}\label{conclusion}

In this survey, we have proposed a categorization for reinforcement learning exploration strategies based on the information that the agent uses in its exploratory action selection. We divided exploration techniques into two general classes: ``Reward-Free Exploration'' and ``Reward-Based Exploration''. The \emph{reward-free} techniques either select actions totally at random, or utilize some notion of intrinsic information in order to  guide exploration, without taking into account extrinsic rewards. This can be useful in environments where the reward signal is very sparse, and therefore not immediately available to the agent.  \emph{Reward-based} exploration methods leverage the information related to the reward signal and can further be divided into the ``memory-free'', which only take into account the state of the environment, and ``memory-based", which consider additional information about the history of the agent's interaction with the environment. In each of these categories, methods which share similar properties are clustered together in one subcategory. The basis for this clustering is mainly the type of information used, as well as the way it is used in the selection of exploratory actions. We have discussed these exploration methods and have pointed out their strengths and limitations, as well as improvements that have been made and some which are still possible. We would like to emphasize that our goal was not to review the theoretical sample complexity results, which are abundant in the field. Rather, we wanted to provide a big picture which captures the current ``lay of the land in terms of methods", and which is useful for practitioners in their choice of methods. We note that theoretical results often need to rely on assumptions about the environment and the RL algorithm, for example tabular or linear representations of the value function, or smooth dynamics, which are often not satisfied in practice. Nonetheless, exploration methods can still provide useful empirical results even if their theoretical assumptions are not satisfied.

 In this study, we have limited ourselves to sequential decision making in  MDPs. We have not covered in detail strategies for POMDPs and bandits, although these settings have provided inspiration  for some of methods proposed for the MDP setting. An emerging trend is the study of safe exploration methods, but as relatively little is written on them so far, we have not focused on this topic. Finally, considering the large number of yearly publications in this field, we have excluded some methods that are similar to the major classes of approaches we discussed.

Owing to the improved and more accessible computational power in the recent years, the newly proposed exploration techniques have contributed significantly to the improvement in the exploration-exploitation dilemma. However, there are major concerns and issues that have not been resolved yet, mainly due to the absence of a consensus over the ways exploration methods can be assessed. For instance, different techniques are evaluated according to different measures of efficiency and performance, such as state coverage, information gain, sample efficiency, or regret. Furthermore, there is no set of standard experimental tasks that all proposed exploration techniques are evaluated on. This diversity in the methods of assessment complicates the comparison of exploration techniques together. Finally, there is often some sort of a discrepancy between the theoretical guarantees that a method provides and the experimental condition the agent encounters.  Consequently, there is often no reliable guarantee for the performance of these techniques in more involved environments. Addressing these issues could be the focus of future work.

\section{Acknowledgement}\label{sec:acknowledge}
The authors would like to thank Scott Fujimoto for providing valuable feedback on the early draft of this manuscript. Funding is provided by Natural Sciences and Engineering Research Council of Canada (NSERC).




\newpage

\newpage

\bibliography{biblio}

\end{document}